\pdfoutput=1
\documentclass[10pt, logo, twocolumn, copyright]{nv}

\usepackage{graphicx}
\definecolor{nvidiagreen}{HTML}{76B900}

\usepackage{amsmath}
\usepackage{amssymb}
\usepackage{mathtools}
\usepackage{amsthm}
\usepackage{multirow}
\usepackage{algorithm}
\usepackage{algpseudocode}
\usepackage{url}            %
\usepackage{booktabs}       %
\usepackage{amsfonts}       %
\usepackage{nicefrac}       %
\usepackage{microtype}      %
\usepackage{xcolor}         %
\usepackage{mathtools}
\usepackage{listings}

\usepackage{xcolor}
\definecolor{pearDark}{RGB}{34,139,34}  
\definecolor{mygreen}{RGB}{34,139,34}
\definecolor{mylightblue}{RGB}{0,162,230}
\definecolor{deepyellow}{RGB}{255,215,0}
\definecolor{nvgreen}{RGB}{118, 185, 0}

\usepackage{xspace}

\usepackage{mdframed}
\usepackage{color}
\usepackage{xcolor}
\usepackage[utf8]{inputenc} 
\usepackage[T1]{fontenc}    

\usepackage{amsfonts}       
\usepackage{nicefrac}       
\usepackage{microtype}      
\usepackage{multirow}
\usepackage{multicol}
\usepackage{tabto}
\usepackage{xspace}
\usepackage{amsmath}
\usepackage{adjustbox}
\usepackage{enumitem}
\usepackage{wrapfig}
\usepackage{times}
\usepackage{verbatim}
\usepackage{amssymb}
\usepackage{mathtools}
\usepackage{caption}
\usepackage{subcaption}
\usepackage{array}
\usepackage{colortbl}
\usepackage{booktabs}
\usepackage{bbm}
\usepackage{makecell}
\usepackage{float}
\usepackage{siunitx}
\usepackage{pifont}
\usepackage{marvosym}
\usepackage{listings}
\usepackage{pdflscape}
\usepackage{footmisc}
\usepackage{url}
\usepackage{tabularx}
\usepackage{arydshln}
\usepackage{hhline}
\usepackage{diagbox}
\usepackage{tcolorbox}
\usepackage[nameinlink]{cleveref}
\usepackage{hyperref}
\usepackage[square,sort,comma,numbers]{natbib} 
\usepackage{fp} 
\usepackage{authblk}
\usepackage{xspace}
\usepackage{xcolor}         
\usepackage{stfloats}
\definecolor{DeepRed}{RGB}{150,20,20}

\crefname{section}{Sec.}{Sec.}
\crefname{proposition}{Proposition.}{Proposition.}
\crefname{equation}{Eq.}{Eqs.}
\crefname{figure}{Fig.}{Figs.}
\crefname{table}{Tab.}{Tabs.}
\crefname{algorithm}{Algorithm}{Algorithms}
\crefname{appendix}{Appendix}{Appendices}
\Crefname{thm}{Thm}{Thm}
\definecolor{codebg}{RGB}{245, 245, 245}
\definecolor{keywordcolor}{RGB}{0, 0, 153}
\definecolor{commentcolor}{RGB}{34, 139, 34}
\definecolor{stringcolor}{RGB}{163, 21, 21}
\definecolor{numbercolor}{RGB}{128, 128, 128}

\title{LongLive-2.0: An NVFP4 Parallel Infrastructure \\ for Long Video Generation}

\runningtitle{LongLive-2.0: An NVFP4 Parallel Infrastructure for Long Video Generation}

\correspondingauthor={
$^{*}$ Equal contribution $^{\dagger}$ Project Lead
}
\author{%
\begin{minipage}[t]{\textwidth}
\centering
Yukang Chen$^{*\dagger}$\quad 
Luozhou Wang$^{*}$\quad 
Wei Huang$^{*}$\quad 
Shuai Yang$^{*}$\quad 
Bohan Zhang\quad \protect\\[0.1cm]
Yicheng Xiao\quad 
Ruihang Chu\quad 
Weian Mao\quad 
Qixin Hu\quad
Shaoteng Liu\quad 
Yuyang Zhao\quad  \protect\\[0.1cm]
Huizi Mao\quad 
Ying-Cong Chen\quad 
Enze Xie\quad
Xiaojuan Qi\quad 
Song Han\protect\\[0.2cm]
NVIDIA \protect\\[0.2cm]
\href{https://github.com/NVlabs/LongLive}{\texttt{github.com/NVlabs/LongLive}}
\end{minipage}
}

\begin{abstract}\small
\textbf{Abstract:} We present LongLive-2.0, 
an NVFP4-based parallel infrastructure throughout the full training and inference workflow of long video generation, addressing speed and memory bottlenecks. 
(1) For training, we introduce sequence-parallel autoregressive (AR) training, instantiated as Balanced SP, which co-designs the efficient teacher-forcing layout with SP execution by pairing clean-history and noisy-target temporal chunks on each rank, enabling a natural teacher-forcing mask with SP-aware chunked VAE encoding.
Combined with NVFP4 precision, it reduces GPU memory cost and accelerates GEMM computation during training, the proportion of which increases as video length grows.
Moreover, we show that a high-quality infrastructure and dataset enable a remarkably clean training pipeline. Unlike existing Self-Forcing series methods that rely on ODE initialization and subsequent distribution matching distillation (DMD), LongLive-2.0 directly tunes a diffusion model into a long, multi-shot, interactive auto-regressive (AR) diffusion model. It can be further converted to real-time generation (4 to 2 denoising steps) with standalone LoRA weights. 
(2) For inference on Blackwell GPUs, we enable W4A4 NVFP4 inference, quantize KV cache into NVFP4 for memory savings, and boost end-to-end throughput with asynchronous streaming VAE decoding. On non-Blackwell GPU architectures, we deploy SP inference to match the speed on Blackwell GPUs, while the quantized KV cache can lower inter-GPU communication of SP. Experiments show up to 2.15$\times$ speedup in training, and 1.84$\times$ in inference. LongLive-2.0-5B achieves 45.7 FPS inference while attaining strong performance on benchmarks. To our knowledge, LongLive-2.0 is the first NVFP4 training and inference system for long video generation.
\end{abstract}

\newcommand{\llteaserfigure}{%
  \begin{center}
    \includegraphics[width=1.0\textwidth]{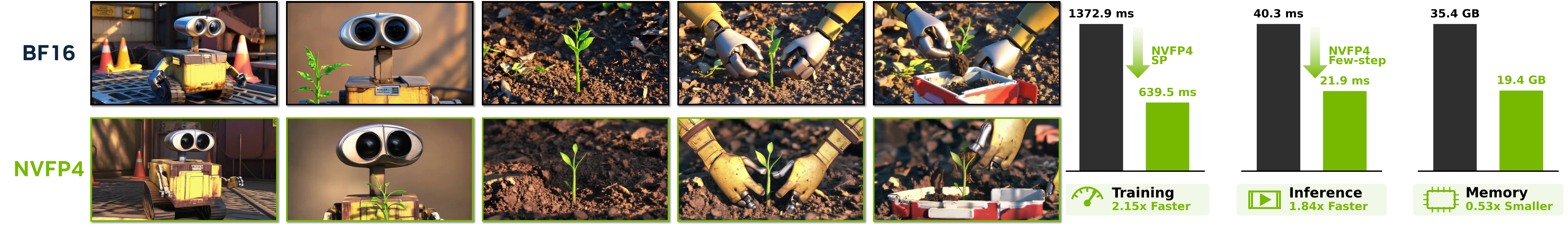}
    \vskip -2pt%
    \captionof{figure}{\textbf{LongLive 2.0 supports NVFP4-based multi-shot long-video generation for both training and inference.}
    {Representative frames from five shots generated with BF16 and NVFP4 (Left):} NVFP4 preserves the overall scene composition, subject structure, and shot-level semantics of the BF16 baseline. Note that LongLive 2.0 allows flexible customization of the duration for each shot.
    {Efficiency and memory comparison (Right):} NVFP4 achieves 2.15$\times$ faster training and 1.84$\times$ faster inference, reducing training latency from 1372.9 ms per iteration to 639.5 ms per iteration and inference latency from 40.3 to 21.9 ms/frame, \textit{i.e.}, 45.7 FPS, while reducing memory usage from 35.4 GB to 19.4 GB.
    }\label{fig:teaser}
  \end{center}%
}

\makeatletter
\renewcommand{\maketitle}{%
  \twocolumn[%
  \begin{adjustwidth}{0pt}{24pt}
    \begin{center}
      {\titlefont \@title\par}%
      \vskip11pt
      {\@author\par}%
      \vskip20pt%
    \end{center}
  \end{adjustwidth}
  \abscontent
  \vskip12pt%
  \llteaserfigure
  \vskip10pt%
  ]%
  \thispagestyle{firststyle}%
  \markboth{\@runningtitle}{\@runningtitle}%
}
\makeatother

\begin{document}
\maketitle

\section{Introduction}
Long video generation suffers from excessive GPU memory consumption and low computational efficiency in both training and inference. For training, a high-quality long video model requires extensive training over massive long-video datasets, leading to prohibitively high computational costs. For inference, long video models are commonly required in interactive and real-time applications that demand strict low latency; yet, the video length poses severe challenges to deployment. Existing works on long video generation primarily focus on algorithmic designs, while largely neglecting infrastructure optimizations for training, inference, and real-world deployment.
\begin{figure*}[t]
\centerline{\includegraphics[width=1.0\textwidth]{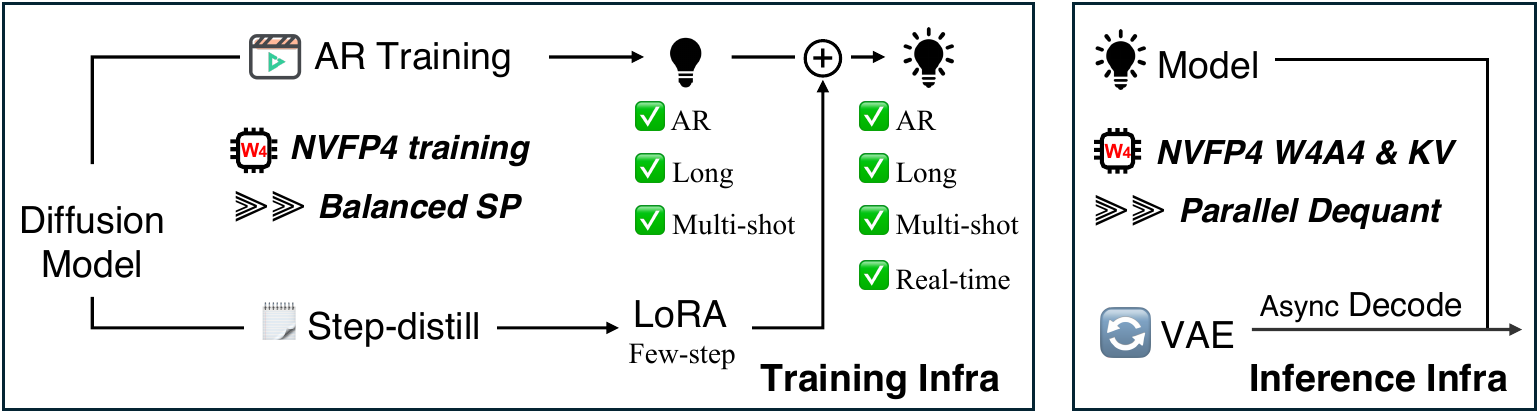}}
\caption{\textbf{Overview of the LongLive-2.0 Framework.} \textbf{Training Infra (Left):} The diffusion model is fine-tuned via AR training on long videos, where Balanced SP and NVFP4 quantization improve training efficiency. In parallel, we derive standalone LoRA weights via DMD training. \textbf{Inference Infra (Right):} Full NVFP4 enables low-precision inference (W4A4) and KV-cache compression. Furthermore, asynchronous decoding eliminates idle time to maximize generation throughput.}
\label{fig:main}
\end{figure*}

Existing works on long video generation still have notable limitations.
At the infrastructure level, few works explore joint co-design between training and inference. For inference, quantization-based methods only adopt post-training quantization (PTQ)~\cite{sageattention,sageattention2,sageattention3}, leading to misalignment between training and inference with suboptimal performance.
At the algorithm level, prevailing training pipelines such as Self-Forcing~\cite{huang2025self} and Causal-Forcing~\cite{zhu2026causal} are overly complicated. Long-video diffusion training typically requires ODE initialization, distribution matching distillation (DMD), and subsequent long tuning in a multi-stage manner.

In this work, we present LongLive-2.0, an NVFP4-based parallel infrastructure for long video generation training and inference, as shown in Figure~\ref{fig:main}. 
On the training side, we introduce sequence-parallel AR training to scale AR training for long videos, with Balanced SP as the current instantiation. Unlike traditional SP, which treats the clean-context and noisy-target latent streams as an ordinary concatenated sequence, Balanced SP assigns each GPU the clean and noisy latents from the same temporal chunk. This paired layout balances loss-bearing tokens across GPUs and enables a natural teacher-forcing~\cite{zhou2025taming} attention mask after Ulysses All-to-All communication. Balanced SP also allows SP-aware chunked VAE encoding so that latent preparation is partitioned consistently with the DiT sequence. Combined with NVFP4 quantization, the training process becomes more memory- and compute-efficient. This efficiency gain becomes increasingly important as input videos grow longer, since both latent preparation and GEMM-heavy DiT computation become increasingly costly. 

On the inference side, Blackwell GPUs allow full NVFP4 alignment between training and inference for highly efficient W4A4 inference, and we further quantize the KV cache into NVFP4 for substantial memory savings. On other GPU architectures (non-Blackwell), SP inference also enables real-time generation; we defer the details to Appendix~\ref{ap:sp_inference}, where the quantized KV cache also lowers inter-GPU communication.
Moreover, LongLive-2.0 targets end-to-end generation speed, a more practical metric than diffusion-model FPS alone. While existing reports often exclude VAE decoding, we reduce this gap with two system-level optimizations: customized parallel dequantization in the NVFP4 KV-cache kernel minimizes the overhead of low-bit KV computation, and asynchronous streaming decoding overlaps VAE decoding with model denoising. As video length increases, decoding overhead is increasingly amortized, allowing end-to-end FPS to approach model-only FPS.

\begin{figure*}[t]
\centerline{\includegraphics[width=1.0\textwidth]{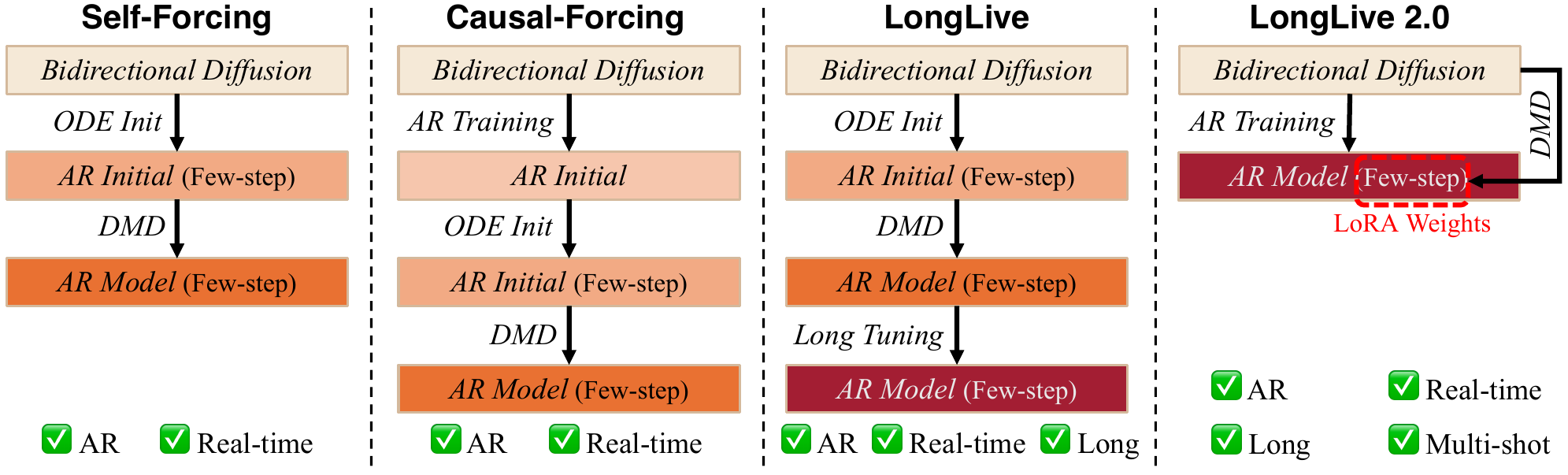}}
\caption{\textbf{Clean Pipeline for AR Video Generation.} LongLive-2.0 bypasses the complex, multi-stage processes (\textit{e.g.}, ODE initialization, intermediate DMD) required by previous methods. Instead, our first stage directly performs AR training on the base bidirectional model using long-video data. By simply injecting standalone LoRA weights to enable few-step inference, we achieve a streamlined pipeline that uniquely supports long, interactive, multi-shot, and real-time generation all at once.}
\label{fig:clean-pipeline}
\end{figure*}

Strong infrastructure can further improve algorithm design. In our case, high-quality training infrastructure enables training models on long videos directly and efficiently, leading to a cleaner pipeline. As shown in Figure~\ref{fig:clean-pipeline}, existing methods~\cite{huang2025self, zhu2026causal} rely on complex multi-stage processes, involving ODE initialization and DMD, but still have limitations in long, interactive, or multi-shot generation. The original LongLive~\cite{yang2025longlive} adds a long tuning stage to support long and interactive generation, but this further complicates the training pipeline. In contrast, LongLive-2.0 directly achieves a long, interactive, multi-shot AR model via long-video fine-tuning. 
The model can then be converted to real-time generation (from 4 to 2 denoising steps) with standalone LoRA weights. Through algorithm–infrastructure co-design, LongLive-2.0 achieves strong performance on video generation benchmarks, including VBench~\cite{vbench} and VBench-Long~\cite{vbench-long}.

\section{Training Infrastructure}

LongLive-2.0 supports a clean training pipeline. We directly fine-tune a bidirectional diffusion model into a long, interactive, multi-shot AR model with long-video data. Meanwhile, we derive standalone LoRA weights via DMD training directly on the original diffusion model. 
With LoRA weights integrated, our AR model seamlessly gains few-step denoising ability and enables real-time inference.

\begin{figure*}[t]
\centerline{\includegraphics[width=1.0\textwidth]{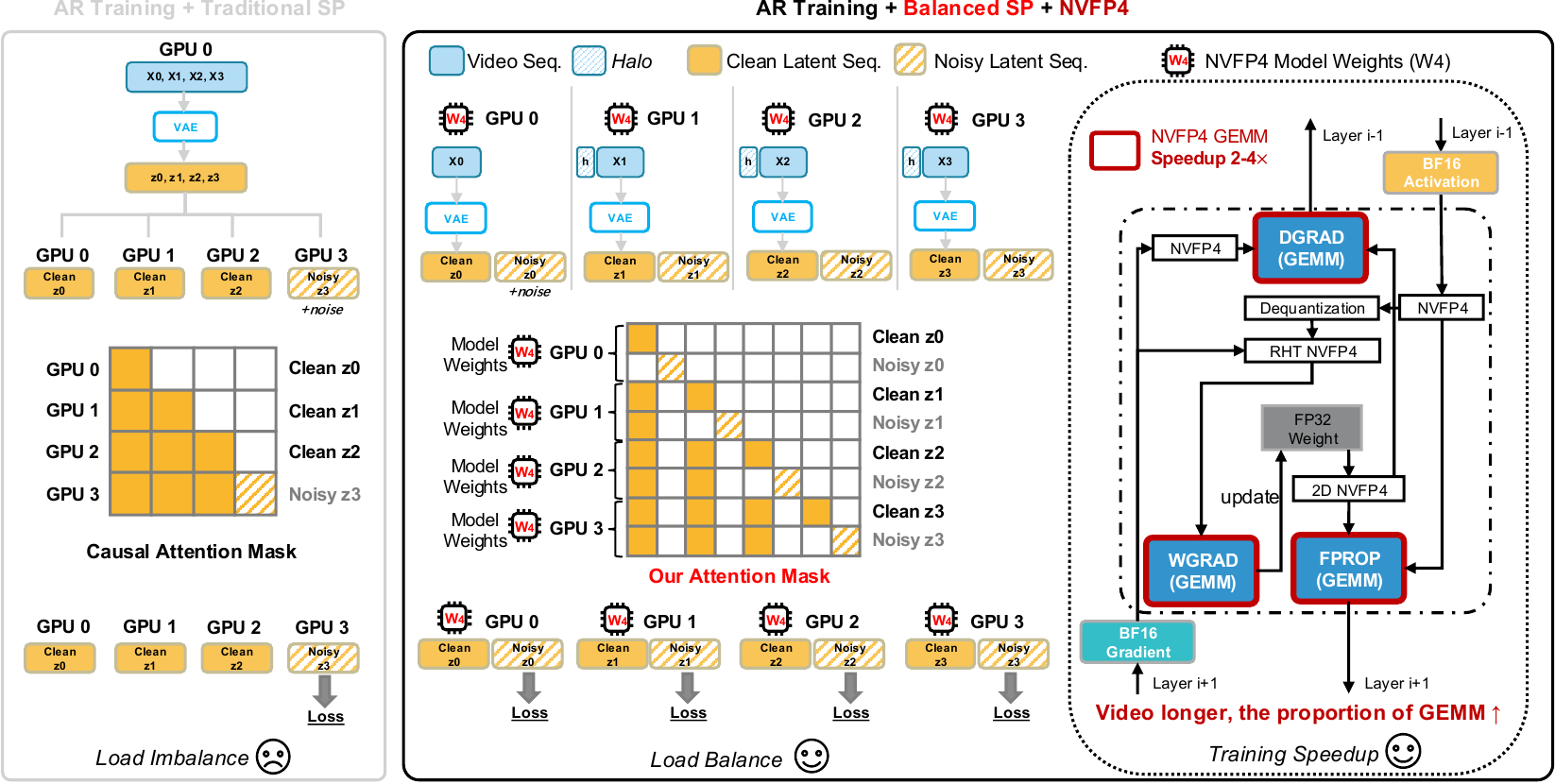}}
\caption{\textbf{Overview of the Training Infrastructure.} \textbf{Traditional SP (Left):} In the efficient teacher-forcing layout, clean history tokens and noisy target tokens are concatenated into one sequence. Naive SP treats this as a general sequence, causing loss computation workload imbalance and leaving VAE encoding replicated across sequence-parallel ranks. 
\textbf{Balanced SP (Middle):} The same temporal chunk ownership is reused across clean/noisy latent streams, SP attention, VAE encoding, and loss computation. This chunk-aligned layout balances loss-bearing tokens across GPUs while avoiding replicated VAE preparation.
\textbf{NVFP4 (Right):} NVFP4 training orthogonally accelerates GEMMs and reduces memory footprint.}
\label{fig:sp-training}
\end{figure*}

\subsection{Sequence-Parallel AR Training}\label{sec:sp_training}
LongLive-2.0 trains a chunk-level AR diffusion model that denoises the current noisy chunk conditioned on clean generated history.
We use clean-context teacher forcing~\cite{jin2024pyramidal, li2024autoregressive, zhou2025taming, zhang2025test, zhang2025generative} rather than diffusion forcing~\cite{chen2024diffusion} to avoid the train-test gap, but a literal teacher-forcing pass supervises only one target suffix at a time.
Following the efficient parallel teacher-forcing formulation summarized in Self-Forcing~\cite{huang2025self}, for an $N$-chunk raw video window $\mathbf{X}$ we encode the raw video into VAE latents $\mathbf{Z}$ and form paired streams $[\mathbf{z}_{clean};\,\mathbf{z}_{noisy}]$.
A block-sparse AR mask lets each noisy chunk attend to preceding clean chunks and its own noisy tokens, so one forward pass supervises all $N$ noisy chunks.

This efficient formulation makes the AR objective practical, but it also creates a structured long sequence that quickly exceeds the memory capacity of a single GPU.
Naively applying SP to AR video training leaves two inefficiencies.
First, slicing the concatenated DiT sequence $[\mathbf{z}_{clean};\,\mathbf{z}_{noisy}]$ can create clean-heavy and noisy-heavy ranks, which imbalances the loss-bearing workload.
Second, the VAE stage still encodes the full video on every SP rank (or on one root rank followed by broadcast), so latent preparation does not benefit from sequence sharding.
We therefore co-design the AR training layout with the sequence-parallel data layout and instantiate it as Balanced SP on top of DeepSpeed-Ulysses~\cite{jacobs2023deepspeed}.
Balanced SP shares the same temporal partition across VAE preparation, local clean/noisy latent construction, DiT attention, and loss computation; under this layout, the block-sparse AR attention mask is generated directly on the SP-native token order.
Balanced SP constructs the paired clean/noisy streams locally on each rank.
Rather than materializing a full $[\mathbf{z}_{clean};\,\mathbf{z}_{noisy}]$ sequence on one rank and then slicing it, rank $p$ prepares its own clean latent chunk and applies the noise schedule locally to obtain the matched noisy chunk.
Using $\mathbf{z}$ to denote the DiT sequence after patch embedding, let $P$ be SP group size, $L$ be the total clean-plus-noisy token length, $H$ be the number of attention heads, and $d$ be the head dimension. Rank $p$ owns
\begin{equation}
    \mathbf{z}^{(p)} = \left[\mathbf{z}_{clean}^{(p)},\;
    \mathbf{z}_{noisy}^{(p)}\right]
    \in \mathbb{R}^{\frac{L}{P} \times H \times d}.
\end{equation}
This paired layout gives every rank both context and target tokens from the same temporal range, making the loss computation uniform across ranks.

The same chunk ownership is also applied before the DiT.
Each rank VAE-encodes only its local raw-video chunk $\mathbf{X}^{(p)}$ plus a left halo that covers the VAE temporal receptive field, then discards the halo latents and keeps the exact local latent chunk $\mathbf{Z}^{(p)}$.
If $F$ is the number of latent frames and $h$ is the halo size, replicated VAE encoding costs $O(F)$ per rank, while Balanced SP reduces the per-rank VAE cost to $O(F/P+h)$ without changing the DiT training objective.
After Ulysses All-to-All, the paired layout naturally produces an interleaved global token order.
Rather than materializing a permutation back to $[\mathrm{all\ clean};\,\mathrm{all\ noisy}]$ at every attention layer, we construct the AR mask directly on this communication-native order and compile it with \texttt{flex\_attention}~\cite{dong2024flex}.
Appendix~\ref{ap:balanced_sp} gives the exact halo construction, natural-mask index mapping, global-coordinate handling, and SP-sharded error-buffer design.

\subsection{NVFP4 Training}
NVFP4~\cite{nvidia2024blackwell} is attractive for long-video generation, 
because it reduces memory cost and accelerates low-precision GEMMs, whose share grows as video length increases.
We therefore use NVFP4 for both AR training and DMD step distillation. To the best of our knowledge, this is the first end-to-end NVFP4 recipe for long video generation.

\textbf{NVFP4 Preliminaries.} $\;$
NVFP4 represents each tensor element using a 4-bit floating-point value in the E2M1~\cite{ocp2023mx} format together with hierarchical scaling~\cite{abecassis2025pretraining,cook2025four}. For a tensor $\mathbf{X}$, the dequantized tensor can be written as:
\begin{equation}
    \hat{\mathbf{X}} = \hat{\mathbf{X}}^{\text{FP4}} \cdot \alpha^{\text{FP8}} \cdot \alpha^{\text{FP32}}, \qquad \hat{\mathbf{X}}^{\text{FP4}} \in \mathbb{F}_{\mathrm{E2M1}},
\end{equation}
where $\alpha^{\text{FP8}}$ is a block-wise (16 elements) scale stored in FP8 E4M3 and $\alpha^{\text{FP32}}$ is a tensor-wise global scale stored in FP32. For a tensor $\mathbf{X}$, we set:
\begin{equation}
\begin{aligned}
\hat{\mathbf{X}}
&=
\hat{\mathbf{X}}^{\mathrm{FP4}}
\cdot
\alpha^{\mathrm{FP8}}
\cdot
\alpha^{\mathrm{FP32}}, \\
\hat{\mathbf{X}}^{\mathrm{FP4}}
&\in
\mathbb{F}_{\mathrm{E2M1}}.
\end{aligned}
\end{equation}

where $B_i$ denotes the $i$-th 16-element quantization block, $M^{\text{FP8}} = 448$ is the maximum representable magnitude of E4M3, and $M^{\text{FP4}} = 6$ is the maximum representable magnitude of E2M1. Unlike uniform integer quantization, FP4 uses non-uniform dynamic step sizes, providing finer resolution for small values and coarser spacing for large ones. In addition, NVFP4 is natively supported on NVIDIA Blackwell GPUs, enabling more efficient hardware acceleration for low-precision computation.

\textbf{Multi-Shot AR NVFP4 Training.} $\;$
In AR training, we train the AR long-video generator on real multi-shot data with the AR objective described in \S~\ref{sec:sp_training} and the multi-shot prompting interface in \S~\ref{sec:interactive_training}, using end-to-end NVFP4 quantization. At the 5B scale, this requires custom quantization and dequantization kernels together with dedicated CUDA kernels for NVFP4 GEMMs; for the RHT-enabled branch, we additionally use Triton kernels for the transformed quantization and dequantization path. As shown in Figure~\ref{fig:sp-training}, we apply the standard NVFP4 recipe to the linear layers: 2D block scaling for weights, 1D block scaling for activations and gradients, and higher precision for numerically sensitive operations such as reductions, normalization statistics, and optimizer states. This follows prior NVFP4 training practice and preserves consistency across forward and backward GEMMs~\cite{abecassis2025pretraining,castro2025quartet}. For the most gradient-sensitive path, we use prior stabilization techniques, notably Random Hadamard Transform (RHT) before quantization on the operands of the weight-gradient GEMM. In our 64s training setting, this NVFP4 stack provides an approximately $1.8\times$ training speedup.

\begin{figure}[t]
\centerline{\includegraphics{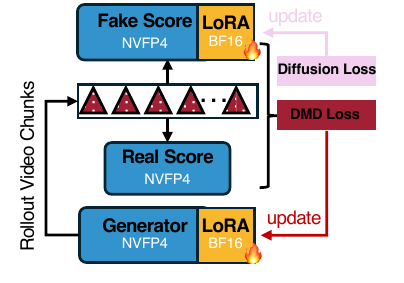}}
\caption{\textbf{NVFP4 DMD training infrastructure.} The generator, real-score model, and fake-score model are colocated under a low-precision NVFP4 setup.}
\label{fig:dmd_training}
\vspace{-10pt}
\end{figure}

\textbf{Few-step Distillation in NVFP4.} $\;$
In few-step distillation, both teacher and student operate in W4A4 NVFP4, keeping distillation tightly aligned with inference. As shown in Figure~\ref{fig:dmd_training}, the \textit{Real-Score} model is quantized to W4A4 for NVFP4 inference. We use adaptive block scaling via scale search~\cite{cook2025four} to quantize NVFP4 weights and activations: besides the standard target magnitude 6, the quantizer also evaluates 4 and selects the lower-error encoding for each block (Appendix~\ref{ap:4o6}). This adaptive search reduces weight quantization error under W4A4 inference. The trainable \textit{Fake-Score} model and \textit{Generator} use the same W4A4 NVFP4 backbone, freeze the quantized backbone, and optimize only LoRA adapters:
\begin{equation}
\begin{aligned}
\mathbf{W}
&\simeq
\operatorname{Dequant}\!\left(Q_{\mathrm{search}}(\mathbf{W}_0)\right)
+ \Delta \mathbf{W}, \\
\Delta \mathbf{W}
&=
\frac{\alpha_{\mathrm{LoRA}}}{r}\mathbf{B}\mathbf{A}.
\end{aligned}
\end{equation}
where $\mathbf{W}_0$ is the pretrained backbone weight, $Q_{\mathrm{search}}$ denotes scale-search-based NVFP4 quantization, $\mathbf{A}$ and $\mathbf{B}$ are trainable low-rank matrices of rank $r$, and $\alpha_{\mathrm{LoRA}}$ is the LoRA scaling factor. Restricting updates to a LoRA subspace follows recent low-bit adapter tuning in LLMs~\cite{dettmers2023qlora,huang2025qerl} and is more stable in our DMD setting than updating the full quantized backbone~\cite{yang2025longlive,zhu2026causal,huang2025self}. The DMD objective is unchanged (\S~\ref{sec:dmd}); only the LoRA weights are trainable.

\section{Inference Infrastructure}
\subsection{NVFP4 Inference}\label{sec:nvfp4_inference}
At deployment time, we execute the generator in W4A4 NVFP4, either as a quantized backbone with a separate LoRA branch or as a merged W4A4+LoRA model with fused low-rank kernels. Since AR long-video generation is dominated by repeated linear layers and attention GEMMs, replacing BF16 GEMMs with FP4 GEMMs reduces memory traffic and offers an ideal theoretical throughput speedup of up to $4\times$. We additionally materialize quantized weights and drop BF16 master weights after LoRA wrapping, further reducing resident memory. Unlike post-training quantization (PTQ) methods~\cite{zandieh2025turboquant,li2024svdquant,zhao2024vidit}, our backbone is trained with NVFP4-aware training, which better preserves generation quality under W4A4 inference.

\subsection{Parallel KV Quantization}
\begin{figure}[t]
\centerline{\includegraphics[width=\columnwidth]{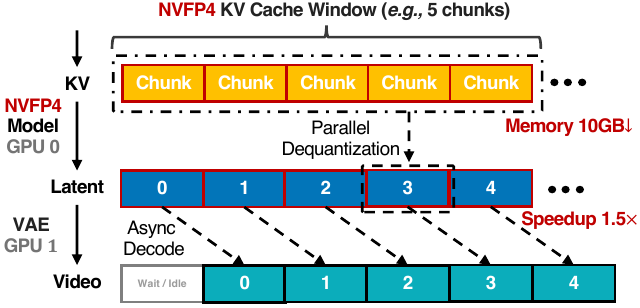}}
\caption{\textbf{NVFP4 inference infrastructure.} LongLive-2.0 combines W4A4 NVFP4 inference, quantized KV cache,  and asynchronous VAE decoding to improve throughput and reduce memory for long-video generation.}
\vspace{-10pt}
\label{fig:async_inference}
\end{figure}

In AR long video generation, KV cache memory grows linearly with history and quickly becomes a bottleneck~\cite{xi2026quant}. We therefore quantize the cache at the frame-chunk level, aligned with our blockwise pipeline. Each chunk contains $F_c=8$ frames and $T_c = F_c L_f$ latent tokens. For layer $\ell$, the cached KV chunk $c$ is
\begin{equation}
\mathbf{K}_{\ell,c}, \mathbf{V}_{\ell,c} \in \mathbb{R}^{T_c \times H \times d},
\end{equation}
which we reshape to $\mathbb{R}^{(T_c H)\times d}$ and quantize independently with NVFP4 micro-block scaling. For keys, we first apply a simple $K$-smoothing:
\begin{equation}
\bar{\mathbf{K}}_{\ell,c}[t,h,:]
=
\mathbf{K}_{\ell,c}[t,h,:]
- \frac{1}{d}\sum_{u=1}^{d}\mathbf{K}_{\ell,c}[t,h,u].
\end{equation}
We then apply the same adaptive scale selection described in Equation~\ref{eq:4o6}, without repeating the notation here. The storage cost changes from $4 T_c H d \quad \text{bytes}$ to $\frac{9}{8} T_c H d \quad \text{bytes}$, ignoring the amortized tensor-wise scale and padding overhead, which is close to a $3.6\times$ KV-cache compression ratio in practice. This chunkwise NVFP4 cache preserves generation quality while substantially reducing memory footprint. Since LongLive-2.0 uses sink-token sliding windows, each attention step may access multiple cached chunks; we therefore implement a customized parallel CUDA dequantization kernel for efficient in-window reconstruction (Figure~\ref{fig:async_inference}). This keeps the overall KV-cache quantization/dequantization overhead below $2\%$ in practice.

\subsection{Asynchronous Streaming Decoding}
The final variational autoencoder (VAE) decoding step is often a major bottleneck in video generation. The centralized decoding scheme used in the baseline LongLive model accumulates all latent chunks before sequential decoding, leading to a VAE-side GPU memory cost of $\mathcal{O}(C \cdot T_c)$ for $C$ chunks and a long end-to-end latency.
We instead design a heterogeneous asynchronous pipeline. We first re-engineer the 3D VAE to support chunk-by-chunk streaming decoding with immediate CPU offloading, reducing the VAE GPU memory footprint to $\mathcal{O}(T_c)$. We then dedicate one GPU to VAE decoding and run it asynchronously alongside the $P$-GPU DiT SP cluster. Let $t_{\text{DiT}}$ and $t_{\text{VAE}}$ denote the per-chunk latencies of denoising and decoding, respectively. 
While the DiT cluster denoises chunk $c+1$, the VAE node decodes chunk $c$. Since the DiT loop is dominant in practice ($t_{\text{DiT}} \ge t_{\text{VAE}}$), decoding is largely hidden behind denoising, reducing the end-to-end latency from $C(t_{\text{DiT}} + t_{\text{VAE}})$ to approximately $C \cdot t_{\text{DiT}} + t_{\text{VAE}}$ and enabling memory-efficient streaming generation.

\section{Algorithm-level Designs}
\subsection{Training in Clean Pipeline}
\textbf{Multi-Shot Interactive AR Training.}\label{sec:interactive_training}$\;$

The AR objective and efficient teacher-forcing layout are described in \S~\ref{sec:sp_training}; here we focus on the algorithmic interface enabled by chunk-level generation. We employ Wan2.2-TI2V-5B~\cite{wan} as our base model.
We treat each temporal latent chunk $\mathbf{Z}_i$ as an editable generation unit and bind it to an individual text prompt $\mathbf{T}_i$.
Cross-attention is factorized per chunk as $\text{CrossAttn}(\mathbf{Z}_i, \mathbf{T}_i)$, rather than conditioning the whole video on a single global prompt.
This decoupling lets different shots carry different prompts, supports prompt switches at chunk boundaries, and preserves previously generated history when the user edits future chunks.

\textbf{Few-step Distillation.} \label{sec:dmd}$\;$
Our few-step distillation framework is derived from LongLive, but with several important simplifications. 
First, because the AR-trained model already supports long-video generation, we avoid the original multi-stage strategy with ODE initialization, short-video DMD, and streaming long-tuning DMD.
We instead perform one-stage DMD distillation on top of the AR-trained model, yielding a cleaner formulation without separate initialization or progressive long-tuning stages. 
Second, we do not fully fine-tune the DiT backbone; instead, we optimize LoRA modules only during the entire distillation process. This choice leads to more stable optimization and makes the resulting few-step capability easily transferable to any Wan2.2-TI2V-5B-based AR model. 
Specifically, we initialize the student, critic, and teacher from the original Wan2.2-TI2V-5B model. Similar to LCM-LoRA~\cite{luo2023lcm}, we find that the trained LoRA can be directly plugged into the AR model to reduce inference steps without further tuning.
In the end, the distilled model reduces generation to two steps, while preserving the long-video generation ability of the original framework.
We discuss the differences between our strategy and straightforward DMD fine-tuning in Appendix (\S~\ref{ap:dmd_comparison}).

\begin{figure}[t]
\centerline{\includegraphics[width=0.72\columnwidth]{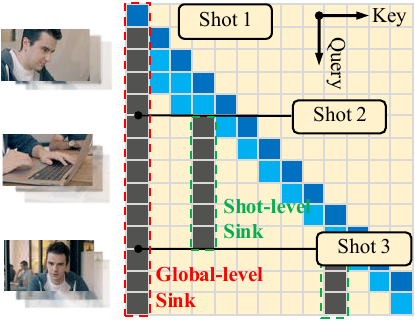}}
\caption{\textbf{Multi-shot Attention Sink for streaming multi-shot inference.}}
\label{fig:shot-level-sink}
\end{figure}
\subsection{Inference with Multi-Shot Attention Sink}\label{sec:multi_shot_attention_sink}

To deploy our model for multi-shot streaming, we adopt sliding-window self-attention with KV caching to cap the per-step compute footprint at $\mathcal{O}(W\!\cdot\! L_c)$, where $W$ is the attention-window length in chunks and $L_c$ is the token length of each chunk. 
However, naively discarding tokens outside the window causes appearance drift. While standard attention sinks~\cite{xiao2023efficient} mitigate this by pinning the first few video frames, they fail in multi-shot settings: a single global sink cannot preserve \textit{intra-shot} coherence, while a moving shot-level sink loses \textit{global} identity.

\begin{table}[t]
\centering
\caption{AR training efficiency of LongLive-2.0. We compare end-to-end iteration time (seconds); red subscripts denote speedup over BF16+SP.}
\label{tab:wan22_ffn1_nvfp4}
\scriptsize
\resizebox{\linewidth}{!}{
\begin{tabular}{c c c c c}
\toprule
\rowcolor{gray!12}
\shortstack{\textbf{Input}\\\textbf{Length}} &
\shortstack{\textbf{BF16}\\\textbf{w/o SP}} &
\shortstack{\textbf{BF16}\\\textbf{w/ SP}} &
\shortstack{\textbf{BF16}\\\textbf{Balanced SP}} &
\shortstack{\textbf{NVFP4}\\\textbf{Balanced SP}} \\
\midrule
16s & 75.3 & 52.2 & 45.8 &
$\mathbf{40.1}_{\textcolor{DeepRed}{\scriptstyle \mathbf{1.3}\times}}$ \\
32s & 202.7 & 162.7 & 136.8 &
$\mathbf{119.3}_{\textcolor{DeepRed}{\scriptstyle \mathbf{1.4}\times}}$ \\
64s & OOM & 1372.9 & 1196.5 &
$\mathbf{639.5}_{\textcolor{DeepRed}{\scriptstyle \mathbf{2.1}\times}}$ \\
\bottomrule
\end{tabular}}
\end{table}
\begin{table}[t]
\centering
\caption{Progressively quantizing the generator, real-score, and fake-score models in DMD training. We report peak per-GPU memory.}
\label{tab:dmd_nvfp4_scaling}
\scriptsize
\setlength{\tabcolsep}{2.8pt}
\renewcommand{\arraystretch}{1.12}
\resizebox{\linewidth}{!}{%
\begin{tabular}{l l l r c}
\toprule
\rowcolor{gray!12}
\textbf{Generator} &
\textbf{Real} &
\textbf{Fake} &
\shortstack{\textbf{Peak Memory}} &
\shortstack{\textbf{Ratio} $\downarrow$} \\
\midrule
BF16 & BF16 & BF16 & 70.5 GB & - \\
\textbf{NVFP4} & BF16 & BF16 & 63.3 GB & 0.90$\times$ \\
\textbf{NVFP4}+LoRA & \textbf{NVFP4} & BF16 & 57.2 GB & 0.81$\times$ \\
\cdashline{1-5}
\textbf{NVFP4}+LoRA &
\textbf{NVFP4} &
\textbf{NVFP4}+LoRA &
\textbf{49.0 GB} &
\textcolor{DeepRed}{\textbf{0.69$\times$}} \\
\bottomrule
\end{tabular}%
}
\end{table}

\textbf{Multi-Shot Attention Sink.} $\;$
To resolve this, we introduce a multi-shot attention sink with two cooperating anchor sets (Figure~\ref{fig:shot-level-sink}): \textit{Global Sink} ($\mathcal{A}_{g}$): the first $S_g$ frames of the video, permanently fixed to preserve global identity. \textit{Shot-Level Sink} ($\mathcal{A}_{s}$): the first $S_s$ frames of the \emph{current} shot, re-bound at every scene cut to maintain local temporal coherence.

At any chunk generation step $t$, the effective key/value set is $\mathcal{K}_{\text{eff}}(t) = \mathcal{A}_{g} \cup \mathcal{A}_{s} \cup \mathrm{KV}_{[t-W, t)}$, with overlapping tokens deduplicated. $\mathcal{A}_{s}$ incurs zero memory overhead: it is tracked merely via two scalar pointers (\textsc{start}, \textsc{len}). It is virtually prepended to the sliding window only after the window rolls past it, avoiding data copying.

\textbf{Interaction with Chunk-wise Prompting.} $\;$
Crucially, this mechanism integrates seamlessly with our chunk-wise interactive prompting (\S~\ref{sec:interactive_training}). A prompt switch $p_{k}\!\to\!p'_{k}$ inherently defines a scene cut. This simply triggers the local re-binding of $\mathcal{A}_{s}$ to the new chunk and re-initializes the subsequent cross-attention cache, leaving the global sink $\mathcal{A}_{g}$ and preceding history untouched. This strict decoupling enables minute-scale interactive generation without redundant recomputation.

\begin{table*}[t]
\centering
\caption{Inference efficiency of LongLive-2.0 under progressively enabled optimizations. The experiments are handled on NVIDIA GB200 180GB GPU, another GPU is used under \textit{Async Decoding}; end-to-end (E2E) latency and peak
memory are reported at different target video lengths.}
\label{tab:inference_progressive}
\scriptsize
\setlength{\tabcolsep}{4.2pt}
\renewcommand{\arraystretch}{1.15}
\resizebox{1\textwidth}{!}{%
\begin{tabular}{l l r r r r r r}
\toprule
\multirow{2}{*}{\raisebox{-5ex}{\shortstack[c]{\textbf{Inference}\\\textbf{Settings}}}} &
\multirow{2}{*}{\raisebox{-3ex}{\shortstack[c]{\textbf{FPS$\uparrow$}}}} &
\multicolumn{2}{c}{\textbf{16 s}} &
\multicolumn{2}{c}{\textbf{32 s}} &
\multicolumn{2}{c}{\textbf{64 s}} \\
\cmidrule(lr){3-4}\cmidrule(lr){5-6}\cmidrule(lr){7-8}
& &
\shortstack{\textbf{E2E Gen.$\downarrow$}\\\textbf{(s)}} &
\shortstack{\textbf{Total Mem.$\downarrow$}\\\textbf{(GB)}} &
\shortstack{\textbf{E2E Gen.$\downarrow$}\\\textbf{(s)}} &
\shortstack{\textbf{Total Mem.$\downarrow$}\\\textbf{(GB)}} &
\shortstack{\textbf{E2E Gen.$\downarrow$}\\\textbf{(s)}} &
\shortstack{\textbf{Total Mem.$\downarrow$}\\\textbf{(GB)}} \\
\midrule
BF16 & 24.8 & 26.6 & 36.4 & 53.2 & 36.4 & 112.9 & 36.4 \\
\cdashline{1-8}
\textbf{NVFP4} & 32.0 & 22.9 & 29.7 & 46.6 & 29.7 & 96.0 & 29.7 \\
+ \textbf{NVFP4 KV Cache} & 29.7 & 23.8 & \textbf{19.4} & 48.9 & \textbf{19.4} & 99.5 & \textbf{19.4} \\
+ \textbf{Async Decoding} & 29.7 &
15.9 & {\textbf{19.4}} &
29.1 & {\textbf{19.4}} &
57.6 & {\textbf{19.4}} \\
\textbf{3 Steps} & 35.2 &
12.7 & {\textbf{19.4}} &
23.2 & {\textbf{19.4}} &
46.0 & {\textbf{19.4}} \\
\textbf{2 Steps} & \textbf{45.7} &
\textbf{11.2} & {\textbf{19.4}} &
\textbf{19.2} & {\textbf{19.4}} &
\textbf{36.3} & {\textbf{19.4}} \\
\bottomrule
\end{tabular}%
}
\end{table*}
\begin{table*}[t]
  \centering
  \caption{Comparison on VBench among LongLive-2.0 and baselines. \#Steps means denoising steps.}
  \label{tab:baseline_fps}
  \resizebox{\textwidth}{!}{
  \begin{tabular}{@{}l c c c c c c c c@{}}
    \toprule
    \multirow{2}{*}{\textbf{Model}} &
    \multirow{2}{*}{\textbf{Precision}} &
    \multirow{2}{*}{\textbf{\#Steps}} &
    \multirow{2}{*}{\textbf{\#Params}} &
    \multirow{2}{*}{\textbf{Resolution}} &
    \textbf{Throughput} &
    \multicolumn{3}{c}{\textbf{Evaluation Scores} $\uparrow$} \\
    & & & & & \textbf{(FPS)} $\uparrow$ & \textbf{Total} & \textbf{Quality} & \textbf{Semantic} \\
    \midrule
    Self-Forcing~\cite{huang2025self}        & BF16 & 4  & 1.3B & $832{\times}480$  & 21.2 & 84.31 & 85.07 & 81.28 \\
    Causal-Forcing~\cite{zhu2026causal}      & BF16 & 4  & 1.3B & $832{\times}480$  & 21.0 & 84.04 & 84.59 & \textbf{81.84} \\
    Rolling-Forcing~\cite{liu2025rolling}    & BF16 & 4  & 1.3B & $832{\times}480$  & 19.5 & 81.22 & 84.08 & 69.78 \\
    Context-Forcing~\cite{chen2026context}   & BF16 & 4  & 1.3B & $832{\times}480$  & 17.0 & 83.44 & 84.98 & 77.29 \\
    CausVid~\cite{yin2024slow}               & BF16 & 4  & 1.3B & $832{\times}480$  & 21.2 & 81.20 & 84.05 & 69.80 \\
    SANA Video-480P~\cite{chen2025sanavideo} & BF16 & 4  & 2B   & $832{\times}480$  & 13.2 & 84.17 & 84.85 & 81.46 \\
    SANA Video-720P~\cite{chen2025sanavideo} & BF16 & 4  & 2B   & $1280{\times}720$ & --   & 84.05 & 84.63 & \underline{81.73} \\
    Wan2.1-T2V-1.3B~\cite{wan}               & BF16 & 50 & 1.3B & $832{\times}480$  & 1.6  & 84.26 & 85.30 & 80.09 \\
    Wan2.2-TI2V-5B~\cite{wan}                & BF16 & 50 & 5B   & $1280{\times}720$ & 3.3  & 83.32 & 84.95 & 76.81 \\
    \midrule 
    LongLive~\cite{yang2025longlive}         & BF16 & 4  & 1.3B & $832{\times}480$  & 20.7 & \underline{84.87} & \textbf{86.97} & 76.47 \\
    \cdashline{1-9}
    \multirow{3}{*}{\textbf{LongLive-2.0}}   & BF16  & 4 & 5B & $1280{\times}720$ & 24.8 & \textbf{85.06} & \underline{86.67} & 78.63 \\
                                             & NVFP4 & 4 & 5B & $1280{\times}720$ & \underline{29.7} & 84.51 & 86.43 & 76.81 \\
                                             & NVFP4 & 2 & 5B & $1280{\times}720$ & \textbf{45.7} & 83.14 & 85.40 & 74.12 \\
    \bottomrule
  \end{tabular}}
\end{table*}
\section{Experimental Results}
\subsection{Training Efficiency}
\textbf{AR Training Efficiency.}$\;$
Table~\ref{tab:wan22_ffn1_nvfp4} reports the end-to-end AR training iteration time under BF16, BF16+SP, BF16+Balanced SP, and NVFP4+Balanced SP. Plain BF16 is efficient only at shorter video lengths, taking 75.3s and 202.7s at 16s and 32s but running out of memory (OOM) at 64s. Adding sequence parallelism makes long-video training feasible and reduces the 16s/32s iteration time to 52.2s and 162.7s, respectively, while our Balanced SP further improves the BF16 path to 45.8s, 136.8s, and 1196.5s across the three lengths.

Combining Balanced SP with NVFP4 gives the fastest training configuration. It reduces the iteration time to 40.1s, 119.3s, and 639.5s for 16s, 32s, and 64s videos, corresponding to 1.3$\times$, 1.4$\times$, and 2.1$\times$ speedups over the BF16+SP baseline. The gain becomes most pronounced at the longest sequence length, where NVFP4+Balanced SP nearly halves the iteration time compared with BF16+Balanced SP and more than doubles throughput over BF16+SP. 

\begin{table*}[t]
\centering
\caption{Comparison on VBench-Long for 60s video generation. Scores are reported in percentage. Avg. Rank is computed over the six metrics. The best is in bold, and the second-best is underlined.}
\setlength{\tabcolsep}{5pt}
\renewcommand{\arraystretch}{1.08}
\resizebox{\textwidth}{!}{
\begin{tabular}{lccccccc}
\toprule
\textbf{Method} &
\textbf{\makecell{Avg.\\Rank $\downarrow$}} &
\textbf{\makecell{Subject\\Consistency $\uparrow$}} &
\textbf{\makecell{Background\\Consistency $\uparrow$}} &
\textbf{\makecell{Motion\\Smoothness $\uparrow$}} &
\textbf{\makecell{Dynamic\\Degree $\uparrow$}} &
\textbf{\makecell{Aesthetic\\Quality $\uparrow$}} &
\textbf{\makecell{Imaging\\Quality $\uparrow$}} \\
\midrule

NOVA~\cite{deng2025nova}
& 8.50
& 77.50
& 88.06
& \underline{98.94}
& 12.00
& 47.53
& 44.97 \\

MAGI-1~\cite{sandai2025magi1}
& 6.67
& 79.46
& 87.76
& \textbf{99.26}
& 56.00
& 52.10
& 54.54 \\

Causal-Forcing~\cite{zhu2026causal}
& 6.50
& 93.52
& 94.12
& 95.74
& \textbf{72.32}
& 51.24
& 62.30 \\

SkyReels-V2~\cite{chen2025skyreelsv2}
& 6.00
& 84.99
& 89.95
& 98.67
& 44.00
& 57.64
& 66.67 \\

Self-Forcing~\cite{huang2025self}
& 5.83
& 95.84
& 95.27
& 98.20
& 51.72
& 56.05
& 62.22 \\

CausVid~\cite{yin2024slow}
& 5.33
& 86.75
& 89.85
& 98.47
& 52.00
& \underline{62.88}
& 67.47 \\

Rolling-Forcing~\cite{liu2025rolling}
& 4.50
& 94.09
& 94.47
& 98.65
& 36.00
& \textbf{63.50}
& \textbf{72.42} \\
\midrule
LongLive~\cite{yang2025longlive}
& 4.17
& 97.13
& 95.89
& 98.61
& 44.56
& 58.17
& \underline{67.56} \\
\textbf{LongLive-2.0}
& \textbf{3.67}
& \underline{97.48}
& \textbf{97.00}
& 98.86
& \underline{60.62}
& 53.68
& 65.51 \\
\textbf{$\;\;\to$ NVFP4}
& \underline{3.83}
& \textbf{97.62}
& \underline{96.97}
& 98.94
& 45.88
& 53.72
& 66.24 \\

\bottomrule
\end{tabular}
}
\label{tab:vbench_long_30s_60s}
\end{table*}

\textbf{NVFP4 DMD Training.} $\;$
We next study few-step DMD training, where the generator, real-score model, and fake-score model are co-located on each GPU. Table~\ref{tab:dmd_nvfp4_scaling} shows a progressive NVFP4 conversion path, with \texttt{NVFP4} for the frozen real-score branch and \texttt{NVFP4+LoRA} for the trainable branches. Peak per-GPU memory decreases monotonically from 70.5 GB to 49.0 GB, corresponding to a 21.5 GB reduction per GPU, or 0.69$\times$ of the BF16 baseline.

\subsection{Inference Efficiency}
Tables~\ref{tab:baseline_fps} and~\ref{tab:inference_progressive}
show that LongLive-2.0 achieves strong throughput with a favorable latency--memory trade-off. The 4-step 5B model reaches 29.7 FPS, surpassing all listed baselines, while the 2-step variant further improves throughput to 45.7 FPS. In the progressive ablation, NVFP4 first improves both speed and memory over BF16, and KV-cache quantization further reduces peak memory from 29.7 GB to 19.4 GB with only a modest latency cost. Asynchronous decoding then lowers E2E latency by overlapping denoising and VAE decoding, and the final 2-step system reaches 36.3s E2E latency for 64s videos while maintaining the 19.4 GB memory footprint.

\subsection{Performance}
\textbf{Short-video generation.} $\;$
We first evaluate LongLive-2.0 on short-video generation using the official VBench prompts with our prompt augmentation, as shown in Table~\ref{tab:baseline_fps}.
LongLive-2.0 achieves the strongest performance at the higher $1280{\times}720$ resolution. 
We evaluate LongLive-2.0 under NVFP4 quantization. Reducing denoising steps further increases speed to 35.2 FPS (3 steps) and 45.7 FPS (2 steps). This shows that NVFP4 with fewer denoising steps enables efficient real-time 720p video generation, achieving up to $2{\times}$ speedup over prior methods.

We note that higher resolution does not always yield higher VBench scores. Since VBench resizes videos and samples frames, results depend on the evaluation protocol. Similar trends appear in Table~\ref{tab:baseline_fps}, where 720p models (\textit{e.g.}, Wan2.2-TI2V-5B, SANA) do not consistently outperform 480p. Thus, slightly lower scores at 720p are expected and do not indicate worse quality.

\textbf{Long Video Generation.} $\;$
We evaluate LongLive-2.0 with 4 denoising steps on long video generation using MovieGenBench prompts and VBench-Long. As shown in Table~\ref{tab:vbench_long_30s_60s}, LongLive-2.0 achieves the best average rank among all compared methods on 60s generation, demonstrating strong overall long-range generation ability. Its advantage is most pronounced in subject and background consistency: the NVFP4 model obtains the best subject consistency score of 97.62, while the BF16 model obtains the best background consistency score of 97.00.

\section{Conclusion}
In this work, we present LongLive-2.0, a algorithm–infrastructure co-design system for efficient long video generation across training and inference. For training, we propose Balanced SP along with NVFP4 quantization. For inference, we quantize both the model (W4A4) and KV cache to NVFP4 and accelerate execution with parallel dequantization.
Benefiting from this strong infrastructure, LongLive-2.0 enables a remarkably clean training pipeline that directly fine-tunes diffusion models to long, multi-shot AR without complex ODE initialization or additional long tuning stages. Real-time generation can be further achieved with lightweight LoRA weights. LongLive-2.0 achieves up to 2.1× training speedup and 1.8× inference speedup, with LongLive-2.0-5B supporting 45.7 FPS while maintaining strong benchmark performance. To our knowledge, LongLive-2.0 is the first end-to-end NVFP4 training and inference system tailored for long video generation.

\textbf{Limitations.}$\;$ The acceleration gain from low-bit quantization is hardware-dependent. NVFP4 inference delivers acceleration only on Blackwell GPUs (\textit{e.g.}, GB200), which are equipped with the latest-generation Tensor Cores and optimized kernels. In contrast, non-Blackwell GPUs, like A100 (Ampere architecture) and H100 (Hopper architecture), lack native hardware support for these optimized kernels. To compensate this limitation, we use SP inference as an alternative solution to boost inference efficiency on non-Blackwell platforms (Section~\ref{ap:sp_inference} in the appendix).

\textbf{Broader Impacts.}$\;$
LongLive-2.0 reduces computational costs and lowers the resource threshold for related research and deployment. It shares the ethical impacts with existing video generation models. The NVFP4 and parallelism infrastructure itself involves no negative social implications.

\clearpage
{
  \bibliography{reference}
  \bibliographystyle{plain}
}
\onecolumn

\clearpage
\appendix
\section{Related Work}
\subsection{Long Video Generation}
Recent video generation research has shifted from short-clip bidirectional diffusion transformers to causal autoregressive (AR) synthesis, where videos are generated frame-by-frame or chunk-by-chunk.
CausVid~\cite{yin2024slow} converts a pretrained bidirectional video diffusion model into a causal AR generator and distills it into a few-step streaming model.
MAGI-1~\cite{sandai2025magi1} scales chunk-level AR generation with nearly constant peak inference cost, while AAPT~\cite{lin2025aapt} explores one-step real-time interactive generation.
These works establish AR video generation as a promising formulation for streaming synthesis, but also expose challenges such as exposure bias, error accumulation, memory growth, and long-range temporal drift.

A major line of work addresses the train--test mismatch in AR video diffusion.
Self-Forcing~\cite{huang2025self} trains models under their own rollout distribution rather than only teacher-forced ground-truth contexts~\cite{zhou2025taming, acdit}.
LongLive~\cite{yang2025longlive}, Self-Forcing++~\cite{cui2025selfforcingminutescalehighqualityvideo}, and Rolling Forcing~\cite{liu2025rolling} extend this idea to real-time long-video generation with causal attention, KV re-cache, attention sinks, long-context tuning, joint denoising, and few-step distillation.
More recent forcing-based methods analyze finer-grained mismatch: Causal Forcing~\cite{zhu2026causal} studies the architectural gap between bidirectional teachers and causal students; Context Forcing~\cite{chen2026context} uses long-context teachers and Slow-Fast Memory to supervise long-context students; HiAR~\cite{zou2026hiar} performs hierarchical denoising so that future blocks are conditioned on contexts at matched noise levels; and Diagonal Distillation~\cite{liu2026diagonal} exploits both temporal chunks and denoising steps to improve streaming distillation.

Another important direction studies long-range memory and efficient cache management.
Since dense attention over all generated frames is infeasible, existing methods rely on sliding windows, KV caches, attention sinks, or compressed memory.
However, naive memory reuse can cause identity drift, temporal repetition, or motion stagnation.
LoL~\cite{cui2026lol}, Deep Forcing~\cite{yi2025deep}, Relax Forcing~\cite{zhao2026relax}, MemRoPE~\cite{kim2026memrope}, VideoSSM~\cite{yu2025videossm}, and Hybrid Forcing~\cite{li2026hybrid} improve long-horizon stability through RoPE stabilization, deep attention sinks, structured KV memory, evolving memory tokens, state-space memory, and hybrid linear/sparse attention.
Complementary system-oriented methods reduce the deployment cost of AR video generation: Quant VideoGen~\cite{xi2026quant} compresses KV cache memory, FlowCache~\cite{ma2026flowcache} introduces chunk-wise caching, SCOPE~\cite{cui2026scope} applies selective computation, and Helios~\cite{yuan2026helios} designs a large AR model for real-time long-video generation.

Finally, training-free horizon extension and interactive generation have also emerged.
FLEX~\cite{li2026flex}, Test-Time Correction~\cite{xiang2026ttc}, FreeLOC~\cite{tian2026freeloc}, and PackForcing~\cite{mao2026packforcing} extend pretrained or short-trained models to longer horizons through positional correction, test-time trajectory calibration, or structured cache partitioning.
Anchor Forcing~\cite{yang2026anchor} targets prompt-switching in streaming diffusion~\cite{shotadapter}, while ShotStream~\cite{luo2026shotstream} extends AR generation to multi-shot interactive storytelling.
Overall, AR long video generation has evolved from simply causalizing diffusion models into a broader problem involving rollout alignment, memory design, positional extrapolation, distillation, and efficient deployment.

\subsection{FP4 Quantization}
Low-bit quantization has become a central tool for reducing the cost of large generative models. A substantial body of work studies post-training quantization (PTQ) and quantization-aware training (QAT) for LLMs and diffusion models~\cite{zandieh2025turboquant,li2024svdquant,zhao2024vidit}. Representative techniques improve robustness by correcting outlier channels, smoothing activation ranges, reconstructing layer outputs, or using low-rank compensation~\cite{frantar2022gptq,lin2024awq,xiao2023smoothquant,li2024svdquant,huang2024mixture,huang2026mc}. These methods are highly effective for deployment compression, but most of them still assume integer-style quantization or focus on PTQ, leaving a mismatch between low-precision inference and the precision regime used during training.

Recent work has therefore moved beyond FP8~\cite{micikevicius2022fp8} toward FP4 floating-point training and inference. FP4 is attractive because it can reduce memory traffic and matrix-multiplication cost more aggressively than FP8, but the E2M1 value set is extremely coarse and requires careful scaling. Block-scaled formats such as MXFP4~\cite{ocp2023mx,rouhani2023microscaling} and NVFP4~\cite{nvidia2024blackwell,alvarez2025nvfp4} address this issue through microscaling factors shared by small groups of values. Compared with MXFP4, NVFP4 uses finer 16-element blocks, FP8 E4M3 block scales, and a tensor-level global scale, which improves local dynamic-range tracking and has been shown to offer a favorable accuracy-efficiency trade-off in large-scale studies~\cite{abecassis2025pretraining,chmiel2025fp4}.

Stable end-to-end FP4 training also depends on algorithmic choices beyond the numeric format. Prior studies show that weights, activations, and gradients must be quantized consistently, while numerically sensitive operations such as reductions, normalization statistics, and optimizer states often remain in higher precision~\cite{abecassis2025pretraining,chmiel2025fp4}. Random Hadamard transforms and rotation-based quantization help disperse block-level outliers, stochastic rounding reduces bias in low-precision gradient updates, and adaptive block-scale selection such as Four Over Six further lowers NVFP4 quantization error~\cite{ashkboos2024quarot,abecassis2025pretraining,chmiel2025fp4,cook2025four}. Complementary low-bit adapter methods show that quantized backbones can be paired with trainable low-rank updates for efficient finetuning and reinforcement learning~\cite{dettmers2023qlora,huang2025qerl}.

However, existing FP4 studies are primarily centered on LLM pretraining, LLM finetuning, or general low-bit inference. Autoregressive long-video generation introduces different system pressures: spatio-temporal sequences are much longer, denoising repeatedly stresses the same GEMM and attention paths, KV caches grow with generated history, and quality is sensitive to any mismatch between training, distillation, and deployment precision. Our work studies NVFP4 in this setting, aiming to jointly align stable training, W4A4 inference, KV-cache compression, and long-video deployment.

\subsection{Sequence Parallelism}
To overcome single-device memory limits, sequence parallelism (SP) distributes long sequences across multiple devices. Existing SP techniques primarily follow two paradigms. Ring-style systems~\cite{li2023sequence, liu2023ring, liu2024startrail, gu2024loongtrain, longvila, longrl} partition sequences into chunks, overlapping point-to-point communication with attention computation. Conversely, DeepSpeed-Ulysses~\cite{jacobs2023deepspeed} partitions along the attention head dimension, utilizing All-to-All communication to gather full sequences. This strictly decouples communication from the core attention arithmetic. At extreme scales, hybrid approaches like USP~\cite{fang2024usp} integrate both methods, using Ulysses intra-node and Ring inter-node.

As generative modeling expands from LLMs to Diffusion Transformers (DiTs)~\cite{peebles2023scalable, ma2024latte}, computational bottlenecks shift toward the massive spatio-temporal sequences inherent in video generation. 
Consequently, recent infrastructures customize fundamental SP paradigms for multi-dimensional data. 
For instance, StreamFusion~\cite{yang2026streamfusion} tailors hybrid SP communication to the unique memory profiles of DiTs. 
At the lowest infrastructure level, Dynamic Sequence Parallelism (DSP)~\cite{zhao2024dsp} re-engineers 1D SP by dynamically switching communication across spatial and temporal axes, reducing All-to-All overhead. 
Furthermore, systems like Megatron Core introduce Dynamic Context Parallelism~\cite{nvidia2026dynamiccp} to optimize sequence sharding and activation memory specifically for variable-length video pre-training.

In AR video training, efficient mask-based teacher forcing introduces a structure that is absent from ordinary long-context modeling: the same temporal chunk appears once as clean context and once as a noisy prediction target, and training relies on complex spatio-temporal masks compiled via FlexAttention~\cite{dong2024flex}.
Simply combining this clean/noisy teacher-forcing layout with an existing SP backend is therefore insufficient.
Ring-style methods are difficult to apply directly because their load-balancing assumptions do not align with these irregular block-sparse masks, while a naive Ulysses partition can separate clean-only and noisy-only ranks and leave VAE latent preparation replicated across SP ranks.
We therefore build on DeepSpeed-Ulysses but co-design the AR training layout with sequence-parallel execution. In the current instantiation, Balanced SP lets each rank locally construct paired clean/noisy latents from its temporal chunk, builds a natural teacher-forcing mask on the post-All-to-All order, and shards VAE encoding with an exact left-halo scheme.
This distinguishes our method from prior SP systems, which mainly optimize communication schedules or activation memory for generic long sequences rather than the clean/noisy pairing and latent-preparation bottleneck specific to teacher-forced video DiT training.

\section{Multi-shot Long-video Dataset}
We curate a large-scale long-video dataset for training LongLive-2.0. We split raw long videos into independent shots and annotate each with structured captions spanning visual, scene, character, action, and cinematography aspects. After completing shot-level captioning, we merge the captions of all segmented shots from the same full-length video and further refine the integrated descriptions to ensure temporal coherence and logical consistency across consecutive frames and scenes.

Subsequently, we conduct rigorous data filtering and quality cleaning to remove low-quality and invalid samples. We remove videos with excessively short shot duration, content containing logos, watermarks or prominent text, footage with severe camera shake, abnormal playback speed such as fast-forward and slow-motion, overexposed or underexposed frames, blurry and out-of-focus visuals, and low-motion clips with frozen frames or only trivial zoom effects. For further quality control, we adopt the MANIQA~\cite{maniqa} metric to evaluate the visual quality of sampled video frames, and the average score is adopted as the overall quality score for each video. Only top-ranked high-quality videos are retained. In the final version, our dataset contains 120K long videos with abundant segmented shots. The videos are evenly distributed in three duration groups: 16–32 seconds, 32–64 seconds, and over 64 seconds, each accounting for one-third of the total data volume.

\section{Balanced SP Details}\label{ap:balanced_sp}

\textbf{Hybrid parallelism and global coordinates.} $\;$
We use a hybrid scheme with $\texttt{world\_size}=\texttt{dp\_size}\times\texttt{sp\_size}$.
Ranks in the same SP group share the same sample and prompt, while only the temporal token dimension is partitioned.
To match non-parallel training, Rotary Position Embedding (RoPE) uses global frame indices and sequence offsets rather than local rank indices.
The attention mask, supervision mask, and loss mask are also evaluated in global coordinates, and the loss is normalized by the global number of valid tokens.
Consequently, the SP implementation preserves the same training objective as the non-parallel formulation while reducing per-rank activation memory.
Inside each DiT attention block, we use $\mathbf{z}$ to denote the hidden sequence corresponding to the clean/noisy latent streams. Let $P$ be the SP group size, $L$ be the total clean-plus-noisy token length, $H$ be the number of attention heads, and $d$ be the head dimension. The Ulysses backend exchanges the sequence and head dimensions:
\begin{equation}
    \mathbf{z}^{(p)} \in \mathbb{R}^{\frac{L}{P} \times H \times d}
    \xrightarrow{\text{All-to-All}}
    \widetilde{\mathbf{z}}^{(p)} \in
    \mathbb{R}^{L \times \frac{H}{P} \times d},
\end{equation}
so that each device computes full-sequence attention over its assigned $H/P$ heads.
A second All-to-All restores the original sequence-sharded layout before the following FFN.

\textbf{Exact SP-aware VAE encoding.} $\;$
In a naive SP pipeline, each rank either encodes the full video independently or waits for a root rank to encode and broadcast the complete latent sequence.
This makes VAE latent preparation scale with the full video length on every rank, even though the following DiT sequence is already sharded.
Balanced SP keeps the pre-DiT clean/noisy construction local to each rank: each rank VAE-encodes only its local raw-video chunk $\mathbf{X}^{(p)}$ plus a left halo that covers the temporal receptive field of the VAE encoder.
After encoding, the rank discards halo latents, keeps only the local latent chunk $\mathbf{Z}^{(p)}$, and locally forms the matched clean/noisy latent streams.
As long as the halo covers the encoder's left temporal dependency, these local latents are identical to those obtained from full-video encoding, while the per-rank VAE cost is reduced from $O(F)$ to $O(F/P+h)$ for $F$ latent frames, SP size $P$, and halo size $h$.

\textbf{Natural teacher-forcing mask.} $\;$
The standard teacher-forcing mask is defined over the logical DiT sequence layout $[\mathbf{z}_{clean};\,\mathbf{z}_{noisy}]$, where all clean chunks are placed before all noisy chunks.
However, after local paired clean/noisy construction and the Ulysses All-to-All exchange, the communication-native global order becomes
\begin{equation}
    [\mathbf{z}_{clean}^{(0)},\mathbf{z}_{noisy}^{(0)},\ldots,
    \mathbf{z}_{clean}^{(P-1)},\mathbf{z}_{noisy}^{(P-1)}].
\end{equation}
A conventional mask would therefore require an explicit permutation that gathers all clean chunks before all noisy chunks, applies attention in that logical order, and scatters the output back to the SP layout at every attention layer.
We instead evaluate the original teacher-forcing visibility rule directly on the interleaved Ulysses order.
Let $L_{\mathrm{loc}}=L/(2P)$ be the number of clean or noisy tokens contributed by each rank.
For a token index $i$ in the interleaved order,
\begin{equation}
    p(i)=\left\lfloor \frac{i}{2L_{\mathrm{loc}}} \right\rfloor,\quad
    r(i)=i\bmod 2L_{\mathrm{loc}},\quad
    t(i)=p(i)L_{\mathrm{loc}} + (r(i)\bmod L_{\mathrm{loc}}).
\end{equation}
Here $p(i)$ is the rank block, $r(i)$ is the within-rank offset, and $t(i)$ is the original temporal position.
The condition $r(i)<L_{\mathrm{loc}}$ identifies clean tokens, while $r(i)\ge L_{\mathrm{loc}}$ identifies noisy tokens.
Thus each interleaved token has a deterministic logical identity, allowing us to define
\begin{equation}
    M_{\mathrm{nat}}(i,j)=M_{\mathrm{TF}}(\pi(i),\pi(j)),
\end{equation}
where $\pi(\cdot)$ denotes the recovered clean/noisy identity and temporal position.
$\pi$ is never materialized on Q/K/V tensors; the block-sparse mask predicate computes it from token indices and \texttt{flex\_attention} compiles the predicate into the fused attention kernel.
This preserves the conventional teacher-forcing visibility while keeping attention in the communication-native SP order.

\textbf{SP-aware error recycling.} $\;$
A pure teacher-forcing setup still leaves residual exposure bias: during training, the clean prefix is drawn from ground truth, while at inference it consists of the model's own rollout.
We therefore maintain an error-recycling buffer of past latent prediction errors and stochastically inject them into $\mathbf{z}_{clean}$ during training~\cite{li2025stable}.
Under Balanced SP, this buffer must follow the same temporal partition as the DiT sequence; otherwise errors from one SP rank would be replayed at positions that are not reachable by another rank.

Concretely, we use a two-dimensional bucket layout indexed by local block position and diffusion timestep.
The position dimension is sharded by SP: if the full sequence has $N_{\mathrm{blk}}$ temporal blocks and SP size $P$, each rank stores only $N_{\mathrm{blk}}/P$ local block positions together with its global block offset.
This preserves the position-dependent nature of rollout errors while reducing per-rank buffer memory.
For context corruption, the clean prefix error is sampled by matching the local position and marginalizing over timestep, since rollout errors accumulate across the denoising trajectory.
For latent and noise corruption, both local position and timestep are matched.

During warming-up, we gather buffer entries across data-parallel ranks with the same SP rank, rather than across the full world group.
This fills each local position bucket faster using different batch samples, while avoiding cross-SP communication whose positions would be invalid for the current rank.
We shard timestep buckets by SP rank and save one buffer checkpoint per SP rank, which keeps checkpoint size bounded and prevents position-bucket misalignment when resuming training.

\begin{figure}[t]
\centering
\begin{minipage}[t]{0.48\columnwidth}
\vspace{0pt}
\centering
    \IfFileExists{figs/sp_1.pdf}
  {\includegraphics[width=\linewidth]{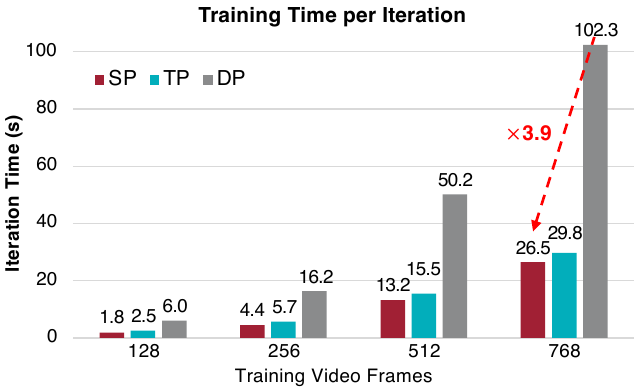}}
  {\fbox{\rule{0pt}{1.38in}\rule{0.95\linewidth}{0pt}}}
\end{minipage}\hfill
\begin{minipage}[t]{0.48\columnwidth}
\vspace{0pt}
\centering
\IfFileExists{figs/sp_2.pdf}
  {\includegraphics[width=\linewidth]{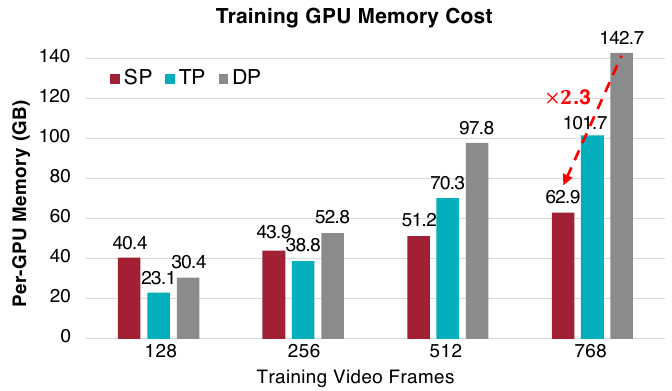}}
  {\fbox{\rule{0pt}{1.38in}\rule{0.95\linewidth}{0pt}}}
\end{minipage}
\caption{Iteration speed and peak memory for sequence parallelism (SP), tensor parallelism (TP), and data parallelism (DP) in interactive AR video generation training on 4 NVIDIA GB200 GPUs. Left: iteration speed. Right: peak memory. SP is fastest at all tested sequence lengths and becomes the most memory-efficient method at long contexts.}
\label{fig:interactive_ar_parallel_scaling}
\end{figure}

Figure~\ref{fig:interactive_ar_parallel_scaling} compares iteration speed and peak memory for sequence parallelism (SP), tensor parallelism (TP), and data parallelism (DP) in interactive AR training with 4 NVIDIA GB200 GPUs.
SP is consistently the fastest, yielding 1.12$\times$--1.41$\times$ speedup over TP and 3.40$\times$--3.86$\times$ over DP.
TP is slightly more memory-efficient at short contexts, but SP becomes the most memory-efficient at long contexts, reducing peak memory to 51.24/62.85 GB at sequence lengths 128/192, compared with 70.26/101.70 GB for TP and 97.75/142.69 GB for DP.
Overall, SP provides the best throughput and memory scaling for long-context interactive AR training.

\section{Sequence Parallelism Inference}\label{ap:sp_inference}

\begin{wrapfigure}{r}{0.60\columnwidth}
\vspace{-5pt}
\centering
\includegraphics[width=\linewidth]{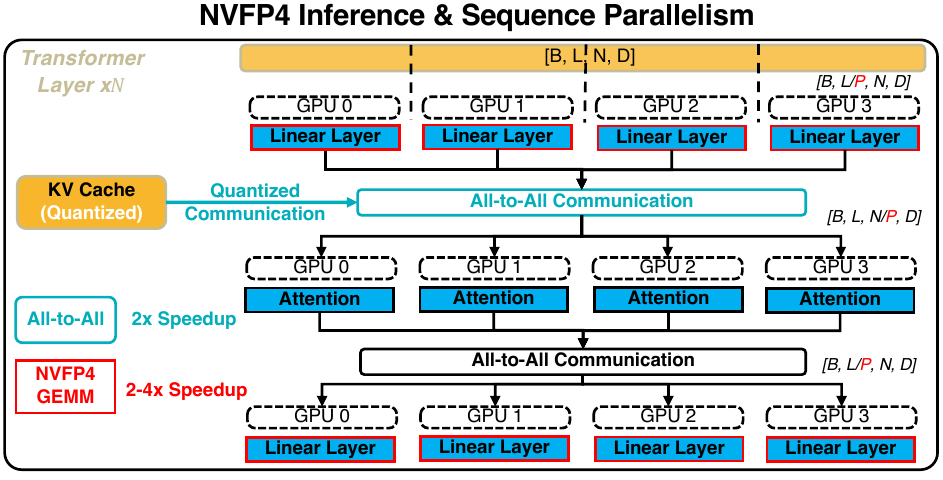}
\caption{\textbf{Sequence Parallelism (SP) Inference.} Inference uses a W4A4 model with SP. KV-cache quantization significantly reduces communication overhead during the All-to-All exchange.}
\label{fig:sp_inference_appendix}
\vspace{-10pt}
\end{wrapfigure}

To reduce memory and latency during extreme long-video generation, we extend the DeepSpeed-Ulysses~\cite{jacobs2023deepspeed} sequence parallelism (SP) strategy from training (\S~\ref{sec:sp_training}) to inference, as illustrated in Figure~\ref{fig:sp_inference_appendix}. Let $P$ denote the SP group size, $L$ the total token sequence length, $H$ the number of attention heads, and $d$ the head dimension. Although SP reduces the per-device memory footprint to $\mathcal{O}(L/P)$, its efficiency is bottlenecked by the \textbf{All-to-All} communication required to transpose the sequence and head dimensions before attention. In a standard BF16 pipeline, exchanging the Query ($\mathbf{Q}$), Key ($\mathbf{K}$), and Value ($\mathbf{V}$) tensors incurs a payload of $\mathcal{O}(L \times H \times d \times 2 \text{ bytes})$ per layer, which heavily stresses interconnect bandwidth.

To mitigate this bottleneck, we combine SP with NVFP4 communication. Since the historical $\mathbf{K}$ and $\mathbf{V}$ tensors are already retrieved from the chunkwise NVFP4 KV cache (\S~\ref{sec:nvfp4_inference}) in compressed form, we also cast the runtime $\mathbf{Q}$ to NVFP4 immediately before the pre-attention All-to-All. Thus, for any tensor $\mathbf{M} \in \{\mathbf{Q}, \mathbf{K}, \mathbf{V}\}$, communication is performed entirely in the low-precision space:
\begin{equation}
\mathbf{M}^{(p)} \in \mathbb{R}^{\frac{L}{P} \times H \times d}_{\text{(NVFP4)}} \xrightarrow{\text{All-to-All}} \widetilde{\mathbf{M}}^{(p)} \in \mathbb{R}^{L \times \frac{H}{P} \times d}_{\text{(NVFP4)}},
\end{equation}
where $p \in \{0, \dots, P-1\}$ is the device index. Executing this transposition natively on NVFP4 data reduces the effective payload from 16 bits to roughly 4.5 bits per element. After accounting for micro-block scaling overhead, the empirical communication volume is reduced by roughly $3.6\times$. This NVFP4-accelerated collective alleviates the bandwidth bottleneck and improves the scalability of SP for long-context AR inference. 
More generally, SP is compatible with a broad range of compression techniques beyond NVFP4 KV-cache quantization, including other low-bit KV compression schemes~\cite{xi2026quant} and attention-pruning methods such as TriAttention~\cite{mao2026triattention}. These methods are complementary to SP: by reducing the tensors exchanged around the pre-attention communication path, they can further lower communication overhead and accelerate inference on non-Blackwell GPUs. We leave a systematic comparison of these alternatives to future work.

\begin{table*}[t]
\centering
\caption{SP inference latency and communication overhead on NVIDIA H100 GPUs. We report end-to-end generation latency and total communication time for BF16 and 4-bit KV-cache settings across different SP group sizes and video lengths. The 64s numbers are estimated from measured shorter-length runs.}
\label{tab:sp_inference_h100}
\scriptsize
\setlength{\tabcolsep}{4.0pt}
\renewcommand{\arraystretch}{1.15}
\begin{tabular}{l l r r r r r r}
\toprule
\rowcolor{gray!12}
& &
\multicolumn{2}{c}{\textbf{16 s}} &
\multicolumn{2}{c}{\textbf{32 s}} &
\multicolumn{2}{c}{\textbf{64 s}} \\
\cmidrule(lr){3-4}\cmidrule(lr){5-6}\cmidrule(lr){7-8}
\rowcolor{gray!12}
\shortstack{\textbf{SP}\\\textbf{Size}} &
\shortstack{\textbf{KV}\\\textbf{Precision}} &
\shortstack{\textbf{E2E Gen.}$\downarrow$\\\textbf{(s)}} &
\shortstack{\textbf{Comm.}$\downarrow$\\\textbf{(s)}} &
\shortstack{\textbf{E2E Gen.}$\downarrow$\\\textbf{(s)}} &
\shortstack{\textbf{Comm.}$\downarrow$\\\textbf{(s)}} &
\shortstack{\textbf{E2E Gen.}$\downarrow$\\\textbf{(s)}} &
\shortstack{\textbf{Comm.}$\downarrow$\\\textbf{(s)}} \\
\midrule
1 & BF16 & 31.0 & --  & 50.2 & --  & 85.0 & --  \\
\cdashline{1-8}
2 & BF16 & 19.3 & 1.8 & 38.1 & 3.2 & 62.5 & 5.4 \\
2 & 4-bit KV Cache & 18.3 & 1.1 & 36.0 & 2.3 & 53.3 & 3.6 \\
\cdashline{1-8}
4 & BF16 & 26.2 & 12.8 & 38.6 & 12.2 & 65.4 & 20.6 \\
4 & 4-bit KV Cache & 21.1 & 7.8 & 32.3 & 9.7 & 54.8 & 16.4 \\
\bottomrule
\end{tabular}
\end{table*}

Table~\ref{tab:sp_inference_h100} verifies that SP inference also provides a practical acceleration path on non-Blackwell GPUs. On H100, moving from single-GPU inference to SP=2 reduces BF16 end-to-end latency from 31.0s/50.2s/85.0s to 19.3s/38.1s/62.5s for 16s/32s/64s videos, respectively. Quantizing the KV cache further reduces the tensors exchanged by SP collectives, cutting communication time from 1.8s to 1.1s for 16s videos at SP=2 and from 12.8s to 7.8s at SP=4. This translates into lower end-to-end latency across the reported lengths, showing that low-bit KV cache compression is an effective way to mitigate the communication overhead introduced by multi-GPU SP inference.

\section{Visual Ablation of Multi-Shot Attention Sink}\label{ap:shot_sink_ablation}

\begin{figure*}[t]
\centering
\IfFileExists{figs/shot-level-sink-ablation.pdf}
  {\includegraphics[width=0.92\textwidth]{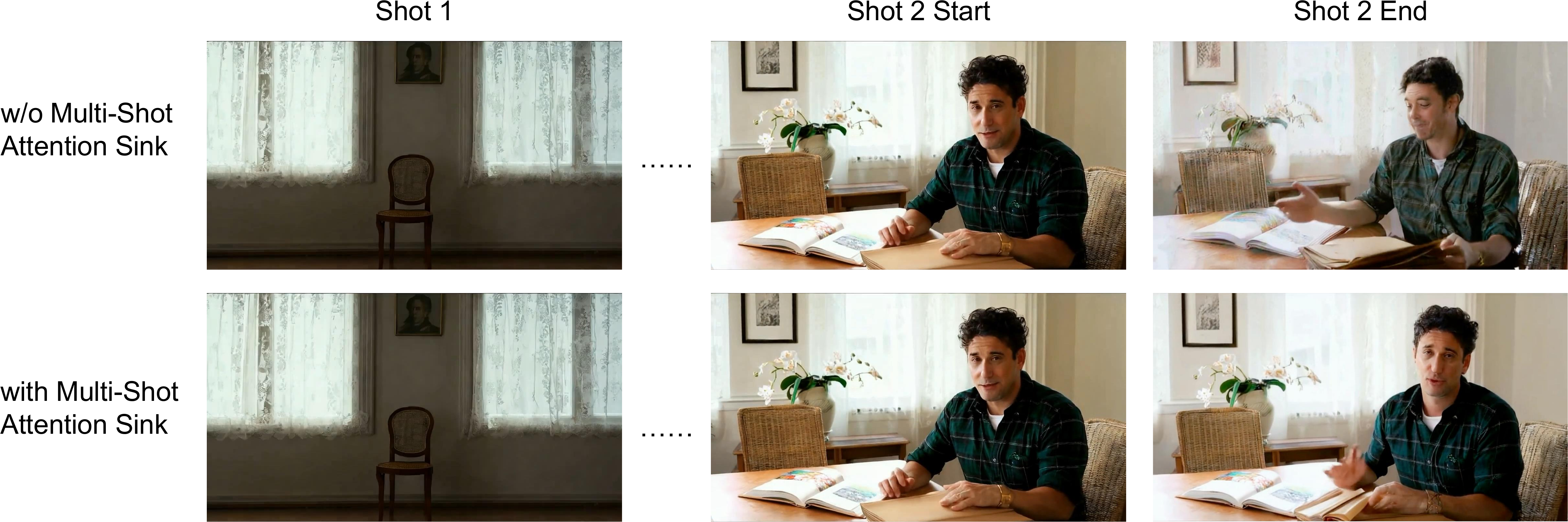}}
  {\fbox{\rule{0pt}{1.85in}\rule{0.92\textwidth}{0pt}}}
\caption{\textbf{Visual ablation of the multi-shot attention sink.} Without the multi-shot attention sink, the generated content drifts. With the multi-shot attention sink stabilizes shot-level appearance.}
\label{fig:shot_level_sink_ablation}
\end{figure*}

Figure~\ref{fig:shot_level_sink_ablation} provides a qualitative ablation of the multi-shot attention sink introduced in \S~\ref{sec:multi_shot_attention_sink}.
Without the multi-shot sink, sliding-window generation can lose shot-local anchors once earlier frames leave the active KV window, causing the later part of a shot to drift in subject appearance and scene layout.
With the proposed multi-shot attention sink, the global sink preserves video-level identity while the shot-level sink keeps the current shot anchored, producing a more stable continuation from the start to the end of the second shot.

\section{Scale Search NVFP4 Quantization}\label{ap:4o6}

The DMD teacher is quantized for W4A4 NVFP4 inference. For teacher weights, we adopt Four Over Six (4/6) adaptive block scaling~\cite{cook2025four}. NVFP4 stores each value in the E2M1 FP4 set
$\{0,\pm0.5,\pm1,\pm1.5,\pm2,\pm3,\pm4,\pm6\}$, together with an E4M3 FP8 scale for every 16-value block and a tensor-level FP32 scale. The standard NVFP4 rule maps the largest absolute value in each block to the largest FP4 magnitude, $6$, which avoids saturation and maximizes dynamic range. However, because FP4 has non-uniform spacing, this choice creates a large representational gap near the block maximum: when the maximum maps to $6$, values between roughly $4/6$ and $1$ of the maximum can only be rounded to $4$ or $6$. As a result, near-maximal values, especially those around $75\%$ of the block maximum, can dominate the quantization error.

Four-Over-Six addresses this issue by also considering a second encoding in which the block maximum is mapped to $4$ rather than $6$. This sacrifices the ability to use the FP4 values $\pm6$ for that block, but makes the high-magnitude region more evenly represented; for example, the FP4 value $3$ then corresponds to $75\%$ of the block maximum. Since this is beneficial only for some blocks and harmful for others, the scale is selected adaptively by explicitly comparing reconstruction error.

Let $\bar{\mathbf{U}}_{B_i} = \mathbf{U}_{B_i} / \alpha^{\text{FP32}}$ denote the globally normalized values in block $B_i$. We define two candidate FP8 block scales:
\begin{equation}
    \alpha^{\text{FP8}}_{i(6)} =
    \operatorname{cast}_{\mathrm{E4M3}}\!\left(
    \frac{\operatorname{max}|\bar{\mathbf{U}}_{B_i}|}{6}
    \right),\qquad
    \alpha^{\text{FP8}}_{i(4)} =
    \operatorname{cast}_{\mathrm{E4M3}}\!\left(
    \frac{\operatorname{max}|\bar{\mathbf{U}}_{B_i}|}{4}
    \right).
\end{equation}
For each candidate scale, the block is quantized to E2M1 FP4 and dequantized back to the original scale. We then select the candidate with lower mean-squared reconstruction error:
\begin{equation}\label{eq:4o6}
\alpha_i^\star=\arg\min_{\alpha \in \{\alpha^{\text{FP8}}_{i(6)}, \alpha^{\text{FP8}}_{i(4)}\}}
\left\|
\mathbf{U}_{B_i} - \hat{\mathbf{U}}_{B_i}(\alpha)
\right\|_2^2,
\end{equation}
where $\hat{\mathbf{U}}_{B_i}(\alpha)$ denotes the dequantized block under block scale $\alpha$. This per-block scale search keeps the standard $6$-based encoding for blocks that require larger dynamic range, while switching to the $4$-based encoding for blocks whose error is dominated by near-maximal values. Because NVFP4 uses E4M3 block scales, the $4$ and $6$ choices can be represented with sufficient fractional precision, enabling this adaptive selection with small quantization-kernel overhead on Blackwell GPUs.

\section{Ablation of NVFP4 Quantization}
\begin{table}[t]
  \centering
  \caption{\textbf{LongLive-2.0 Precision Settings.} We compare BF16 and W4A4 NVFP4 precision under different quantization methods on VBench. \#Step means the number of denoising steps.}
  \label{tab:appendix_ll2_precision_settings}
  \scriptsize
  \begin{tabular}{c c c c c c c c}
    \toprule
    \textbf{Precision} &
    \textbf{Quant.} &
    \textbf{\#Step} &
    \textbf{\#Params} &
    \textbf{Resolution} &
    \textbf{Total$\uparrow$} &
    \textbf{Quality$\uparrow$} &
    \textbf{Semantic$\uparrow$} \\
    \midrule
    BF16 & -- & 4 & 5B & $1280{\times}720$ & 85.06 & 86.67 & 78.63 \\
    NVFP4 & PTQ & 4 & 5B & $1280{\times}720$ & 84.04 & 85.76 & 77.15 \\
    NVFP4 & Pre-trained & 4 & 5B & $1280{\times}720$ & 84.51 & 86.43 & 76.81 \\
    \bottomrule
  \end{tabular}
\end{table}

\begin{figure*}[t]
\centering
\IfFileExists{figs/ptq_nvfp4.pdf}
  {\includegraphics[width=0.98\textwidth]{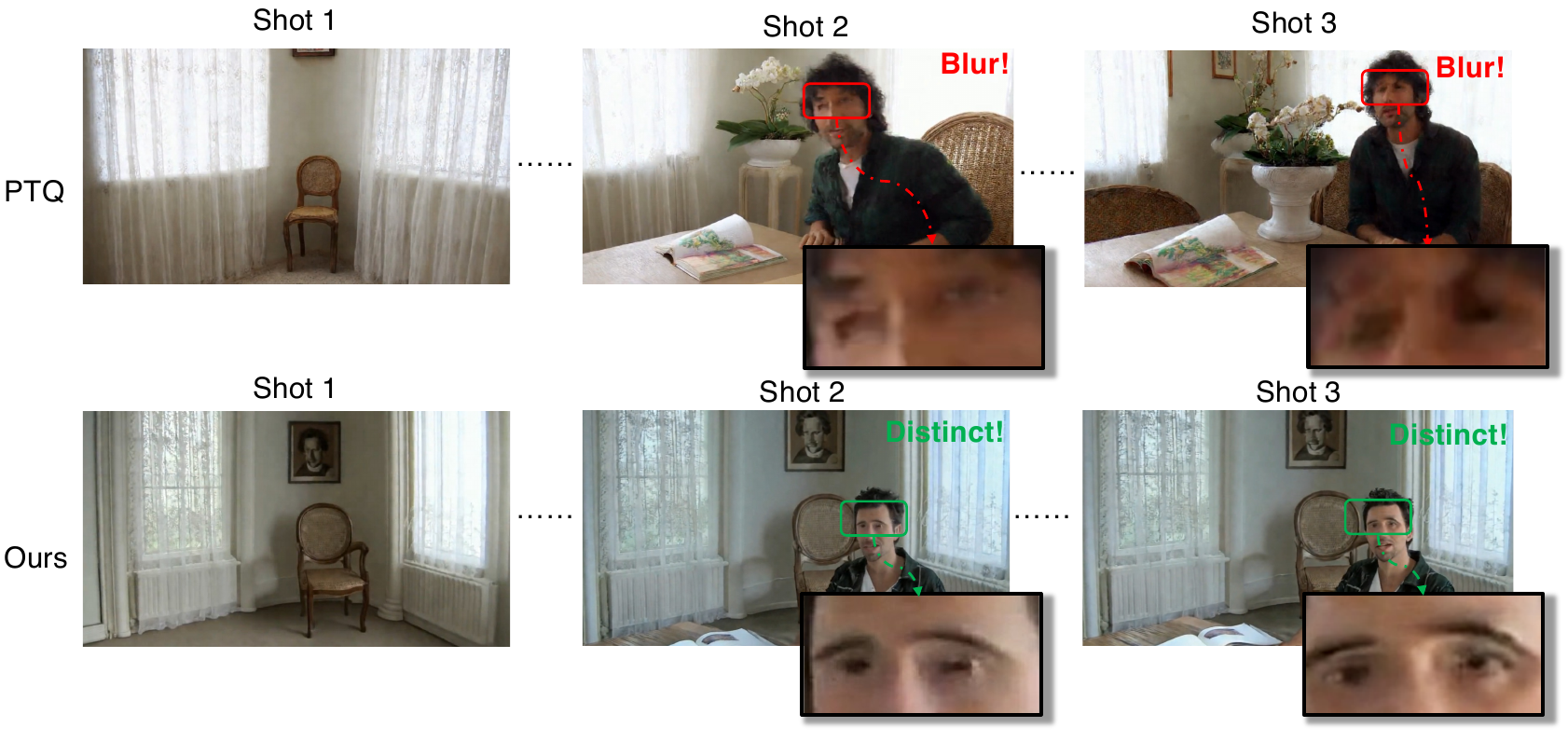}}
  {\fbox{\rule{0pt}{1.85in}\rule{0.92\textwidth}{0pt}}}
\caption{\textbf{Comparison of PTQ and Pre-trained NVFP4.} Top: PTQ. Bottom: pre-trained NVFP4. The first column shows the initial frame, while the following frames compare temporal visual quality. PTQ leads to blurred eyes, whereas pre-trained NVFP4 preserves much clearer facial details.}
\label{fig:ptq_nvfp4_comparison}
\end{figure*}

Figure~\ref{fig:ptq_nvfp4_comparison} provides a qualitative comparison between PTQ and pre-trained NVFP4. The top row shows PTQ results and the bottom row shows pre-trained NVFP4 results; the first column gives the initial frame, and the following frames compare temporal visual quality. PTQ introduces visible degradation, especially blurred eye regions, while pre-trained NVFP4 preserves sharper details.

Table~\ref{tab:appendix_ll2_precision_settings} further isolates the effect of W4A4 NVFP4 quantization on LongLive-2.0 under the same short-video evaluation protocol. The BF16 model serves as the full-precision reference, while direct PTQ converts the trained model to W4A4 NVFP4 only at deployment time. The results show that this direct PTQ path introduces a clear quality drop, indicating a non-negligible mismatch between BF16 training and low-precision W4A4 inference. In contrast, the pre-trained W4A4 NVFP4 setting keeps the model aligned with the target deployment precision and remains close to the BF16 baseline, supporting our design choice of using NVFP4 as a training- and inference-aligned precision rather than only a post-training compression.

\begin{figure*}[t]
\centerline{\includegraphics[width=1.0\textwidth]{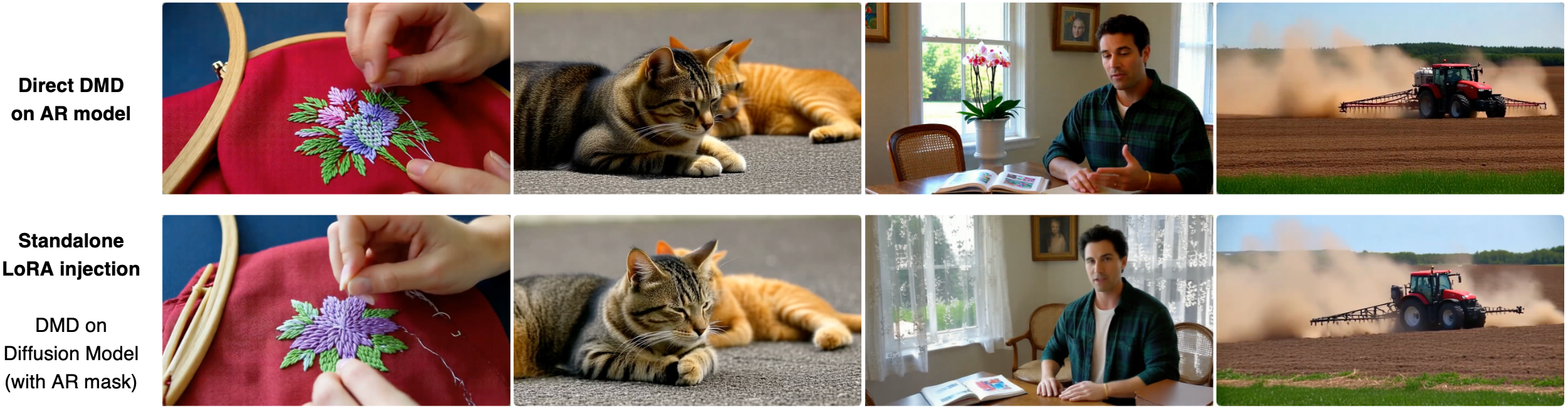}}
\caption{
\textbf{Comparison of two DMD fine-tuning strategies.}
\textbf{(1) Direct DMD fine-tuning of the AR model.}
In this strategy, the student, teacher, and critic are all initialized from the multi-step AR DiT obtained after AR training. This is the most straightforward way to perform DMD fine-tuning, similar to Self-Forcing~\cite{huang2025self}.
\textbf{(2) Standalone LoRA injection.}
In this strategy, the student, critic, and teacher are initialized from the original diffusion model, \emph{e.g.}, Wan2.2-TI2V-5B, while the AR mask is applied to the teacher. DMD fine-tuning is then performed with LoRA.
This strategy is more flexible and convenient: the resulting LoRA weights can be injected into different AR models trained on various types of video data. It also allows DMD fine-tuning to be conducted in parallel with AR training, without waiting for AR training to finish.
As shown in the qualitative comparison, the two strategies lead to different visual characteristics. Direct DMD fine-tuning tends to produce videos with higher contrast and a more synthetic appearance, while standalone LoRA injection yields more natural visual quality. Therefore, we adopt the standalone LoRA injection strategy in our framework.
For a fair comparison, we also use LoRA with the same configuration in the direct DMD fine-tuning setting.
}
\label{fig:dmd_comparison}
\end{figure*}

\section{DMD Training Strategies}\label{ap:dmd_comparison}
We investigate two strategies for DMD fine-tuning to our AR video generation framework, as illustrated in Figure~\ref{fig:dmd_comparison}. 
The first strategy is to directly perform DMD fine-tuning on the AR model. Specifically, the student, teacher, and critic are all initialized from the multi-step AR DiT obtained after AR training. This design is straightforward and follows a similar spirit to Self-Forcing~\cite{huang2025self}. 
The second strategy is standalone LoRA injection, where the student, teacher, and critic are initialized from the original diffusion model, \emph{e.g.}, Wan2.2-TI2V-5B, and the AR mask is applied to the teacher during DMD training. We then train a LoRA module for DMD and inject the LoRA weights into the AR model.

Compared with direct DMD fine-tuning, standalone LoRA injection is more flexible and convenient in practice. Since the LoRA module is trained independently from a specific AR checkpoint, it can be inserted into different AR models trained on various types of video data. Moreover, this strategy allows DMD fine-tuning to be conducted in parallel with AR training, without waiting for the AR training stage to finish. 

Empirically, we also observe different visual characteristics between the two strategies. For a fair comparison, we also apply LoRA with the same configuration in the direct DMD fine-tuning setting. Direct DMD fine-tuning tends to produce videos with higher contrast and a more synthetic appearance, while standalone LoRA injection yields more natural visual quality. 
We note that visual preference can be subjective: the higher-contrast results produced by direct DMD fine-tuning may be appealing in some cases, while we prefer the more natural visual style of standalone LoRA injection and therefore adopt it as our default strategy.

\section{Implementation Details}\label{ap:implementation_details}

We build LongLive-2.0 on Wan2.2-TI2V-5B~\cite{wan}. The text encoder and VAE are kept frozen throughout training. Unless otherwise stated, we use BF16 mixed precision, gradient checkpointing, and AdamW with weight decay $0.01$. For the NVFP4 setting, the GEMM operands in the forward, backward, and weight-gradient paths are quantized to NVFP4, while numerically sensitive operations and optimizer states remain in higher precision.

\textbf{AR training.} $\;$
This stage performs AR training with sequence parallelism. 
We train on $32$ NVIDIA GB200 GPUs with SP size $4$ and hybrid-full FSDP.
The local batch size is $1$ per SP group, with gradient accumulation $2$, giving a global batch size of $16$.
We train for $600$ iterations.
The generator is optimized with learning rate $1.0{\times}10^{-5}$ and AdamW betas $(0.0,0.999)$. We maintain an EMA with decay $0.99$ starting from step $100$.
NVFP4 AR training uses $1920$ NVIDIA GB200 GPU hours.

\textbf{DMD LoRA distillation.} $\;$
This stage distills the AR model with DMD while keeping the pretrained backbone frozen and training LoRA adapters. 
We train on $16$ NVIDIA GB200 GPUs with local batch size $2$ and gradient accumulation $1$, giving a global batch size of $32$.
We train for $5000$ iterations. 
The generator learning rate is $1.0{\times}10^{-5}$ and the critic learning rate is $2.0{\times}10^{-6}$, both with AdamW betas $(0.0,0.999)$. 
The critic is updated every step, and the generator is updated every five steps. 
We use LoRA rank $128$, alpha $128$, dropout $0$, and BF16 adapter weights. 
LoRA is applied to both the generator and fake-score critic, targeting Linear layers inside the causal Wan attention blocks.
NVFP4 DMD LoRA distillation uses $60$ NVIDIA GB200 GPU hours.


\end{document}